\pdfoutput=1

\documentclass[11pt]{article}
\usepackage{amssymb, amsmath, makecell, multirow, hyperref}
\usepackage[table]{xcolor}
\usepackage{booktabs}
 \usepackage{footnote}
\newcommand{\cmark}{\pmb{\checkmark}}
\newcommand{\xmark}{\(\pmb{\times}\)}
\definecolor{lightblue}{HTML}{DCE6F1}

\DeclareMathAlphabet\mathbfcal{OMS}{cmsy}{b}{n}

\newcommand{\mat}[1]{\mathbf{#1}}

\usepackage[final]{acl}

\usepackage{times}
\usepackage{latexsym}

\usepackage[T1]{fontenc}

\usepackage[utf8]{inputenc}

\usepackage{microtype}
\usepackage{inconsolata}
\usepackage{enumitem}

\usepackage{graphicx}

\title{CoLA: Compute-Efficient Pre-Training of LLMs via Low-Rank Activation}

\author{
 \textbf{Ziyue Liu}\textsuperscript{*1}, 
 \textbf{Ruijie Zhang}\textsuperscript{*1}, 
 \textbf{Zhengyang Wang}\textsuperscript{*1}, 
 \textbf{Mingsong Yan}\textsuperscript{1},
 \textbf{Zi Yang}\textsuperscript{2}, \\
 \textbf{Paul Hovland}\textsuperscript{3},
 \textbf{Bogdan Nicolae}\textsuperscript{3}, 
 \textbf{Franck Cappello}\textsuperscript{3}, 
 \textbf{Sui Tang}\textsuperscript{1},
 \textbf{Zheng Zhang}\textsuperscript{1}
\\
\textsuperscript{1}University of California at Santa Barbara;
\textsuperscript{2}University at Albany, SUNY \\
\textsuperscript{3}Argonne National Laboratory \\
\{ziyueliu, ruijiezhang, zhengyangwang, zzhang01\}@ucsb.edu
}

\usepackage{xparse}
\usepackage{amsmath,amssymb,amsfonts,amsthm}

\newtheorem{theorem}{Theorem}[section]

\newtheorem{proposition}[theorem]{Proposition}

\newtheorem{lemma}[theorem]{Lemma}

\newcommand{\norm}[1]{\left\lVert #1 \right\rVert}

\newcommand{\normF}[1]{\left\lVert #1 \right\rVert_{\mathrm{F}}}

\newcommand{\normTwo}[1]{\left\lVert #1 \right\rVert_{\mathrm{2}}}

\newcommand{\bR}{\mathbb{R}}

\newcommand{\rank}{\mathrm{rank}}

\newcommand{\Esigma}{\mathcal{E}_{\sigma}}
\newcommand{\Eone}{\mathcal{E}_{\mathrm{id}}}
\newcommand{\Etwo}{\widetilde{\mathcal{E}_{\mathrm{id}}}}
\newcommand{\Ethree}{\widetilde{\widetilde{\mathcal{E}_{\mathrm{id}}}}}

\newcommand{\col}{\mathrm{col}}
\newcommand{\row}{\mathrm{row}}

\newcommand{\stepjust}[1]{\tag*{\footnotesize\textit{(by #1)}}}
\newcommand{\parens}[1]{\left( #1 \right)}

\newcommand{\din}{d_{\mathrm{in}}}
\newcommand{\dout}{d_{\mathrm{out}}}

\begin{document}
\maketitle
\def\thefootnote{*}\footnotetext{Equal contribution}\def\thefootnote{\arabic{footnote}}
\begin{abstract}
The {\it full-size} MLPs and the projection layers in attention introduce tremendous model sizes of large language models (LLMs),  consuming extensive computational resources in pre-training. We empirically observe that the activations of pre-trained LLMs exhibit low-rank property. Motivated by such observations, we propose {\bf CoLA} and its memory-efficient implementation, {\bf CoLA-M}, to replace these full-size layers with compute-efficient {\bf auto-encoders} that naturally enforce low-rank activations throughout training. This fundamental architectural change eliminates the activation redundancy and significantly boosts model capacity and training efficiency. Experiments on LLaMA models with 60 million to 7 billion parameters show that CoLA reduces the computing cost by $\bf 2\pmb{\times}$ and improves training throughput by $\bf 1.86\pmb{\times}$ while maintaining full-rank level performance. CoLA-M further squeezes memory cost without sacrificing throughput, offering a pre-training approach with collectively superior parameter, computing, and memory efficiency. The LLMs produced are also $\bf 2\pmb{\times}$ smaller, enabling faster inference with lower memory cost on resource-constrained platforms. \footnote{Code available \href{https://github.com/alvin-zyl/CoLA}{here}.}
\end{abstract}

\section{Introduction}
\begin{figure}
\vspace{-10pt}
  \includegraphics[width=\columnwidth]{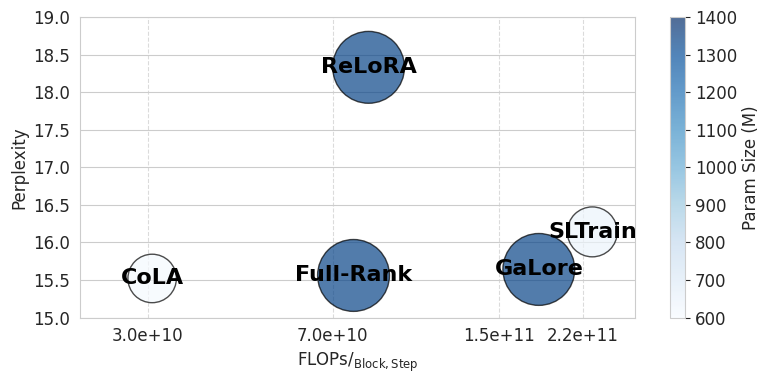}
  \caption{Comparison between various pre-training methods on a LLaMA-1B model with a token batch size of 256. Among them, CoLA is the only one that reduces both compute FLOPs and model size while demonstrating on par validation perplexity with full-rank training.}
  \label{fig:main}
  \vspace{-10pt}
\end{figure}

Large foundation models have achieved unprecedented success in the language, vision, and scientific domains, but they have become huge. Several studies \cite{kaplan2020scaling, hoffmann2022training, krajewski2024scaling, kumar2024scaling} have highlighted a rapid increase in the size of the model and the number of training tokens. Models such as 175B GPT-3 \cite{brown2020language}, 405B LLaMA-3 \cite{dubey2024llama}, and 540B PaLM \cite{chowdhery2023palm} are just a few examples of this trend. 
Under such circumstances, a large number of GPUs are needed in order to provide the computational and high-bandwidth memory capacity needed to pre-train large fundation models over long periods of time (months). This unsustainable trend has prompted the need to develop cost-efficient pre-training techniques that reduce the scale, FLOPs, and GPU memory cost. 

\vspace{1pt}
\noindent\textbf{Motivation:} At the core of increasing resource utilization and cost is the simple practice of scaling up full-size linear layers in decoder-only architectures, which has proven to be a viable and straightforward strategy. Thus, to break free from this unsustainable trend, it is imperative to improve architecture efficiency. This has been widely studied in the deep learning community, involving different levels of factorization of weight matrices: from simple matrix factorizations, i.e., a singular value decomposition (SVD), to higher-order tensor factorizations. Extensive studies have shown that such factorizations can effectively reduce the total number of parameters needed to achieve similar performance in numerous domains \cite{ jaderberg2014speeding, lebedev2014speeding, novikov2015tensorizing, tjandra2017compressing, dao2021pixelated, sui2024elrt, yang2024comera, zhangsparse}, especially when neural networks are overparameterized.

\noindent\textbf{Limitations of state-of-art:} The techniques mentioned above have been applied only to a limited degree to pre-training tasks, and their findings suggest that the pure low-rank or sparse structure often downgrades model performance \cite{khodak2021initialization, kamalakara2022exploring, chekalina2023efficient, zhao2024galore, hu2024accelerating, mozaffari2024slope}. This has pivoted most recent work of efficient pre-training into two directions: 1) Accumulating multiple low-rank updates \cite{huh2024training,lialin2023relora, loeschcke2024loqt}; 2) Enforcing low-rank structures in gradients rather than parameters \cite{zhao2024galore, chen2024fira, huang2024galore, liao2024galore, hao2024flora, zhu2024apollo}. Both approaches have their limitations. 1) The accumulation of low-rank updates requires instantiating a full-rank matrix and a deeply customized training strategy that periodically merges and restarts the low-rank components. This creates computing overhead in practice and can only achieve (if only) marginal computing and memory reduction. 2) Enforcing low-rank gradients reduces only the optimizer memory and adds additional computation that downgrades training throughput. Furthermore, the memory saving caused by gradient compression becomes negligible as the training batch size increases, as activations dominate the total memory cost. Recently SLTrain \cite{han2024sltrain} revisited the notion of parameter efficiency in foundation model pre-training, by having both low-rank factors and an unstructured sparse matrix. SLTrain effectively reduces the total number of parameters without significantly hurting model performance. However, it still introduces computing overhead on top of full-rank training due to the necessary reconstruction of low-rank factors. 
We note that none of the above works has achieved superior efficiency of {\bf parameter}, {\bf computing}, and {\bf memory} simultaneously {\bf without performance drop} in both {\bf training} and {\bf inference} for foundation model pre-training. 

\vspace{5pt}
\noindent\textbf{Contributions:} We rethink the fundamental architecture of LLMs and propose {\bf CoLA}: {\bf Co}mpute-Efficient Pre-Training of LLMs via {\bf L}ow-rank {\bf A}ctivation, and its memory efficient implementation {\bf CoLA-M}, to achieve {\it all} the desirable properties mentioned above. Our contributions include:
\begin{itemize}[leftmargin=*]
\vspace{-5pt}
    \item We propose {\bf CoLA}, a novel architecture to enforce explicit low-rank activations. LLMs use massive full-size MLP and linear layers. CoLA replaces them with auto-encoders. Each auto-encoder applies nonlinear activations between two low-rank factors,  greatly reducing the parameter counts and computing FLOPS while performing on par with the full-rank pre-training.
    
    \vspace{-5pt}
    \item We provide a memory efficient implementation, namely {\bf CoLA-M}, to achieve superior memory reduction without sacrificing throughput.
    \vspace{-5pt}
    \item We theoretically justify the benefit of using CoLA’s auto-encoder structure: they can be strictly better than conventional low-rank models under specific data-dependent conditions. We also derive an effective-rank–aware, non-asymptotic recovery bound that tightens as the spectrum concentrates.
    \vspace{-5pt}
    \item We extensively pre-train LLaMA (with 60M to 7B parameters) and BERT-large. CoLA reduces model size and computing FLOPs by $\bf 2\pmb{\times}$, while maintaining on-par performance to its full-rank counterpart. At the system level, CoLA improves $\bf 1.86\pmb{\times}$ training and $\bf 1.64\pmb{\times}$ inference throughput.  CoLA-M reduces total pre-training memory by $\bf 2/3$, while still manages to improve $\bf 1.3\pmb{\times}$ training throughput over full-rank baselines.
\end{itemize}
A high-level comparison of CoLA(-M) with main baselines is provided in Table~\ref{tab:summary}.

\begin{table}
\centering
\resizebox{\linewidth}{!}{%
\begin{tabular}{c|c|c|c|c|c}
\toprule
\multicolumn{2}{c|}{} & \textbf{CoLA(-M)} & \textbf{SLTrain} & \textbf{GaLore} & \textbf{ReLoRA} \\
\midrule
\multicolumn{2}{c|}{\textbf{Parameter $\pmb{\downarrow}$}} & \cellcolor{green!12}\cmark & \cellcolor{green!12}\cmark & \cellcolor{red!12}\xmark & \cellcolor{red!12}\xmark \\
\midrule
\multirow{2}{*}{\textbf{Compute $\pmb{\downarrow}$}} & Training & \cellcolor{green!12}\cmark & \cellcolor{red!12}\xmark & \cellcolor{red!12}\xmark & \cellcolor{green!12}\cmark \\
& Inference & \cellcolor{green!12}\cmark & \cellcolor{red!12}\xmark & \cellcolor{red!12}\xmark & \cellcolor{red!12}\xmark \\
\midrule
\multirow{2}{*}{\textbf{Memory $\pmb{\downarrow}$}} & Training & \cellcolor{green!12}\cmark & \cellcolor{green!12}\cmark & \cellcolor{green!12}\cmark & \cellcolor{green!12}\cmark \\
& Inference & \cellcolor{green!12}\cmark & \cellcolor{green!12}\cmark & \cellcolor{red!12}\xmark & \cellcolor{red!12}\xmark \\
\midrule
\multirow{2}{*}{\textbf{Throughput $\pmb{\uparrow}$}} & Training & \cellcolor{green!12}\cmark & \cellcolor{red!12}\xmark & \cellcolor{red!12}\xmark & \cellcolor{red!12}\xmark \\
& Inference & \cellcolor{green!12}\cmark & \cellcolor{red!12}\xmark & \cellcolor{red!12}\xmark & \cellcolor{red!12}\xmark \\
\bottomrule
\end{tabular}%
}
\caption{Summary and comparison of different types of efficiency across various pre-training methods.}
\label{tab:summary}
\vspace{-10pt}
\end{table}

\section{Related Work}

\begin{figure*}[t]
\vspace{-10pt}
  \includegraphics[width=\linewidth]{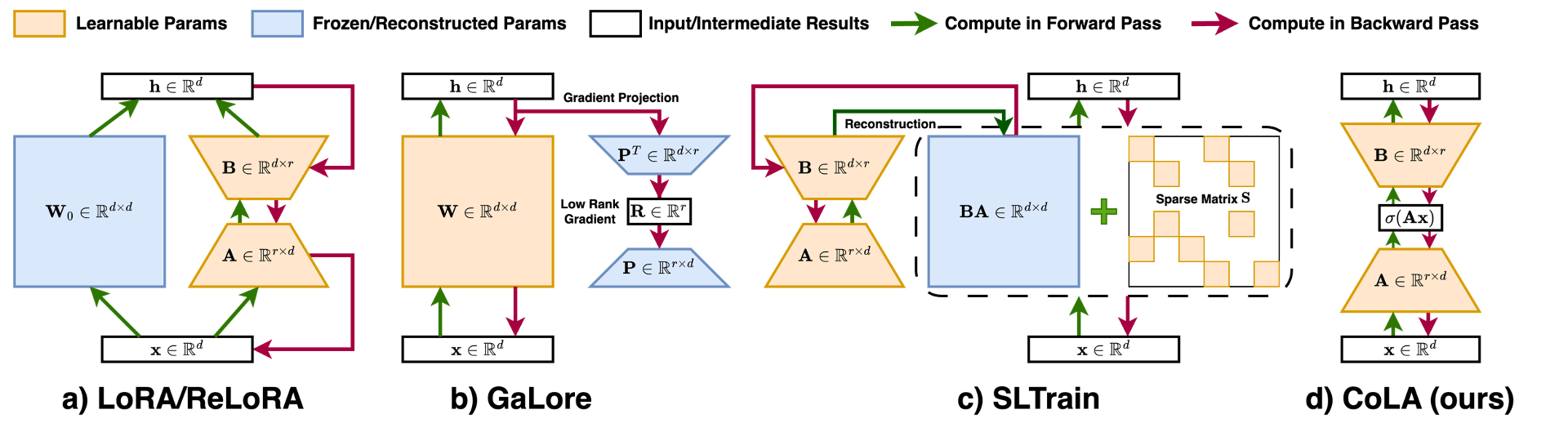}
  \caption{Comparison between different pre-training frameworks. a) LoRA/ReLoRA \cite{lialin2023relora} freezes a full-rank weight; b) GaLore \cite{zhao2024galore} only reduces optimizer states by down and up projecting gradients; c) SLTrain \cite{han2024sltrain} requires reconstruction of the low-rank and sparse matrices; d) CoLA (ours) is a pure low-rank architecture involving only rank $r$ weight matrices.}
  \label{fig:arc}
  \vspace{-10pt}
\end{figure*}

\noindent {\bf Model Compression.} Recent research on efficient LLM pre-training primarily focuses on memory savings. SLTrain \cite{han2024sltrain} is the first method that reduces both trainable parameters and total parameters in LLM pre-training, without significantly hurting model performance. This also reduces memory usage for model, gradients, and optimizer states. However, the existence of its unstructured sparse matrix $\mat{S}$ requires reconstructing $\tilde{\mat{W}} = \mat{BA} + \mat{S}$, otherwise it will incur dense-sparse multiplications that are still memory costly (Fig.~\ref{fig:arc}c). This causes additional computing than the full-rank baseline. LoRA/ReLoRA \cite{hu2021lora, lialin2023relora} reduces trainable parameters by freezing a full-rank $\mat{W}_0$ and training (at least in a later stage) only low-rank factors, potentially reducing memory needs. Yet, any compute savings are limited because the forward pass yields a larger compute than its full-rank counterpart, especially when the rank must stay relatively large in pre-training tasks. LoQT \cite{loeschcke2024loqt} further extends this formulation into quantized training. CoMERA~\cite{yang2024comera} achieves higher model compression and FLOPs reduction, yet its low-rank tensor operations are GPU unfriendly and can also cause a performance drop. Some works investigate pure structured sparsity or combined with low-rank factors \cite{hu2024accelerating, mozaffari2024slope}, but still show a significant performance drop during the pre-training stage.

\vspace{1pt}
\noindent {\bf Gradient Compression.} GaLore \cite{zhao2024galore} reduces memory by projecting gradients into a low-rank space, shrinking optimizer states below the typical $2\times$ AdamW overhead \cite{loshchilov2017decoupled}. However, it adds up/down projections on top of already compute-heavy full-rank training. As shown in Fig.~\ref{fig:main}, its estimated FLOPs surpass full-rank training on the LLaMA-1B scale. Follow-up works \cite{chen2024fira, huang2024galore, liao2024galore, hao2024flora, zhu2024apollo} further explore low-rank gradient projection. While being promising, these methods are mostly orthogonal to our focus. Crucially, they are computing lower-bounded by the full-rank baseline. Our goal instead is to reduce computing cost to a fraction of full-rank LLM pre-training.

\vspace{1pt}
\noindent {\bf Activation Compression.} CompAct~\cite{shamshoum2024compact} reduces memory of the computational graph using low-rank compression on saved activations, which slightly reduces the computing cost of Galore, yet underperforms both GaLore and full-rank training in terms of accuracy. ESPACE~\cite{sakr2024espace} explores a very similar idea by projecting activations based on well-trained weight matrices, thus only applicable to the post-training stage. Crucially, the projections in both methods introduce additional computing costs on top of the full-rank baseline. And both of them do not change the fundamental structure of LLMs.

This paper presents an architectural innovation that explicitly enforces low-rank activations by adopting the {\bf bottleneck-shaped auto-encoders} as the building brick of the transformer architecture. This is conceptually different from the above model compression methods. Our approach is mostly orthogonal with gradient compression techniques, meaning that they could be combined to further boost efficiency.

\section{CoLA for Efficient LLM Pre-Training}
 \begin{savenotes}
\begin{figure}[t]
    \centering
    \includegraphics[width=\linewidth]{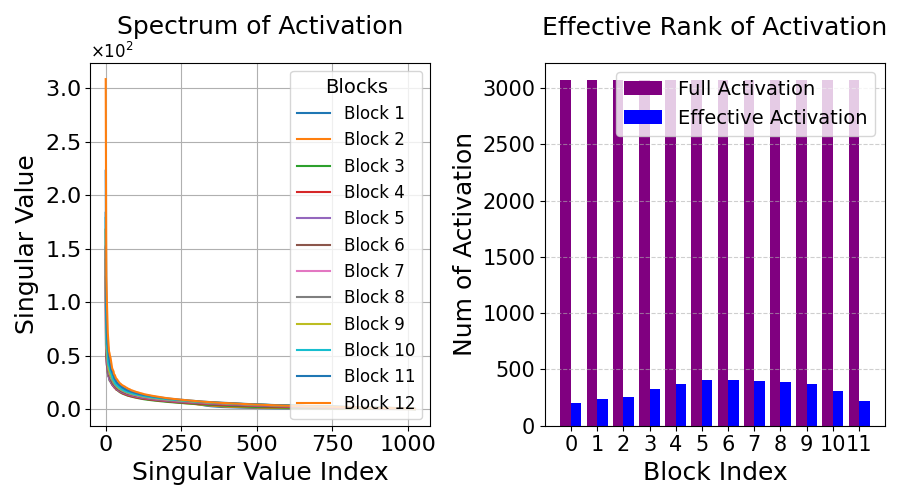}
    \caption{MLP activation [i.e., Eq.\eqref{eq:full-rank-fwd}] spectrum of the pre-trained GPT-2 small \cite{radford2019language}. Model activations are evaluated on the WikiText2 dataset. a) The singular value decay across different decoder blocks. b) The full dimension vs. effective rank ($\alpha=0.95$). \footnote{We updated this figure to reflect the exact post-activation spectrum to avoid potential confusions in our original manuscript.}}
    \label{fig:activation_spectrum}
    \vspace{-10pt}
\end{figure}
 \end{savenotes}

\subsection{A Motivating Example}\label{sec: effective rank}

Many works have observed the low-rank structure of model activations in deep neural networks \cite{cui2020active,huh2021low}. We also observe this phenomenon in LLMs, i.e. the \textit{effective rank} of the activations is much smaller than their original dimensionality. To quantify this, we define the effective rank of a matrix \(\mat{C}\) as the minimal number of singular values needed to preserve an \(\alpha\)-fraction of the total spectral energy. Formally:
\begin{equation}
    r_\alpha (\mat{C}) \;=\; \min \left\{ k \;\middle|\; \frac{\sum_{i=1}^k s_i^2}{\sum_{i=1}^n s_i^2} \;\ge\; \alpha \right\},
\end{equation}
where \(s_1, s_2, \ldots, s_n\) are the singular values of matrix \(\mat{C}\), and \(0 < \alpha \le 1\) is the desired ratio of preserved information. As shown in our experiments, the rapid decay of singular values [Fig.~\ref{fig:activation_spectrum}a] leads to much smaller effective ranks compared to the full dimension [Fig.~\ref{fig:activation_spectrum}b]. This highlights the significant low-rank nature in the activations of pre-trained LLMs. More results showing the same pattern can be found in Appendix~\ref{sec:appendix-low-rank-activation}. 

\subsection{Low-Rank Activation via Auto-Encoder}
\label{subsec:CoLA}

The above observation motivates us to ask one fundamental question: {\it do we really need these full-size MLP and projection layers in LLMs?} To eliminate the redundant activations, we propose to replace them with bottleneck-structured auto-encoders that naturally facilitate low-rank activations.

Let $\mat{W} \in \mathbb{R}^{d_{\text{out}}\times d_{\text{in}}}$ be the weight matrix of an MLP layer: a linear layer followed by a nonlinear activation in the transformer architecture: 
\begin{equation}
    \mat{h}_{\text{MLP}} = \sigma\left(\mat{Wx}\right), \; \text{with} \; \mat{x}\in \mathbb{R}^{d_{\text{in}}}.
    \label{eq:full-rank-fwd}
\end{equation} 
We replace this MLP layer with an auto-encoder layer which consists of low-rank matrices $\mat{A}\in \mathbb{R}^{r \times d_{\text{in}}} $ and $\mat{B}\in \mathbb{R}^{d_{\text{out}} \times r} $ and a non-linear activation $\sigma$ in the middle. Rank $ r<\min(d_{\text{in}, \text{out}}) $ is a design parameter that trades off between compute and performance. Formally, it can be written as:
\begin{equation}
    \mat{h}_{\text{CoLA}} = \mat{B} \, \sigma (\mat{A} \mat{x}).
    \label{eq:main-fwd}
\end{equation}
We empirically find that adding the original nonlinearity on top of Eq.~\eqref{eq:main-fwd} does not harm or necessarily improve the accuracy (c.f. Appendix~\ref{sec:appendix-ablation}). Similarly, for linear layers that are not followed by an activation function, i.e., a projection layer in attention module (we continue using \(\mat{W}\) for simplicity):
\begin{equation}
    \mat{h}_{\text{Linear}} = \mat{Wx},
    \label{eq:full-rank-linear}
\end{equation} 
the low-rank property is also significantly present (see details in Appendix~\ref{sec:appendix-low-rank-activation}). Therefore, they are replaced by CoLA layers \eqref{eq:main-fwd} as well.

The auto-encoder layer naturally enforces a low-rank activation in training, offering a principled approach to eliminate the redundancy observed in Fig.~\ref{fig:activation_spectrum}. We have the following remarks
\begin{itemize}[leftmargin=*]
 \vspace{-5pt}
    \item The auto-encoder layer fundamentally differs from performing low-rank weight compression in an MLP layer. The latter performs lossy compression on model parameters but cannot eliminate the redundancy in activations.

    \vspace{-5pt}
    \item The auto-encoder is not equivalent to using smaller feature dimensions in MLP layers, since $\mat{B}$ in the current layer cannot be merged with $\mat{A}$ in the next layer, due to the existence of various operations (e.g. residual connection, element-wise product) in the original dimension. 
\end{itemize} 

\begin{figure}[t]
\vspace{-10pt}
  \centering
  \includegraphics[width=0.8\columnwidth]{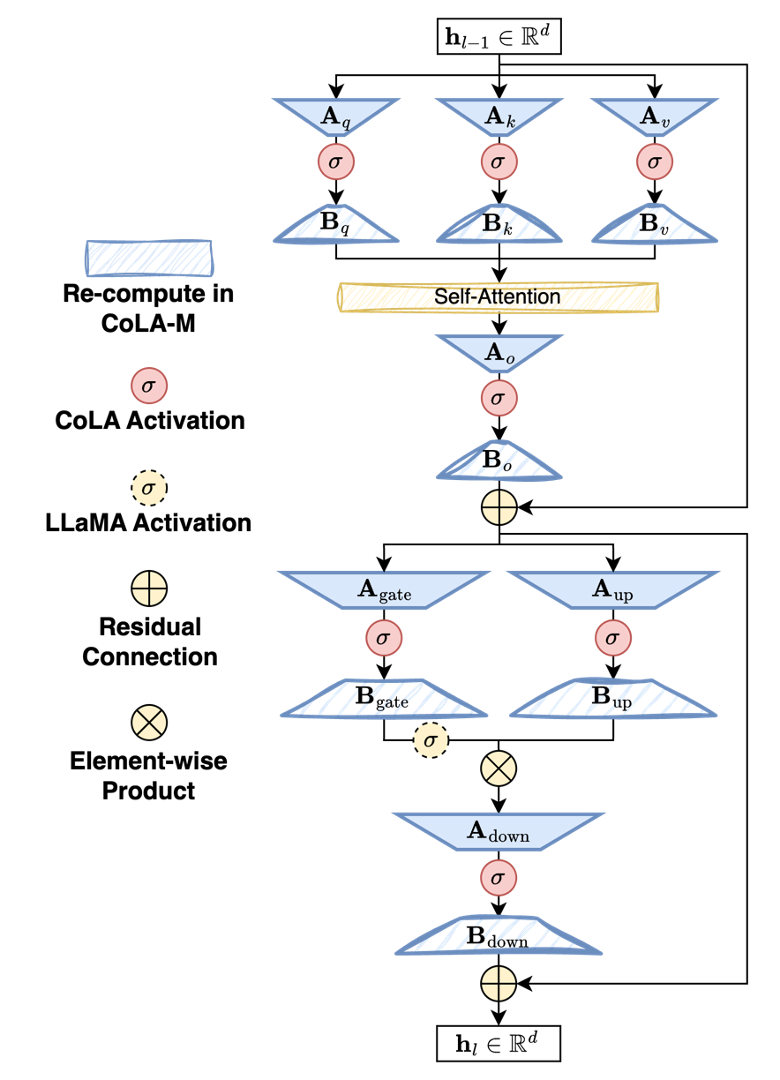}
  \caption{A decoder block in CoLA with LLaMA-like architecture (layer norms, rotary positional embeddings are omitted for simplicity). All MLP layers and projection layers in attention are replaced with auto-encoders. Modules painted in sketch are the re-computations during the backward step of CoLA-M (a memory efficient implementation of CoLA).}
  \label{fig:block}
  \vspace{-10pt}
\end{figure}

Fig.~\ref{fig:block} shows the architecture of each transformer block when adopting CoLA into the LLaMA architecture. We highlight the fact that only the original linear layers and (if any) their follow-up non-linear transformation are modified to the CoLA formulation. Other computations such as the scaled-dot product of the self-attention, as well as residual connections and the element-wise product of LLaMA's MLP layers, remain unchanged.

\subsection{Theoretical Analysis}
We theoretically justify the use of nonlinear activations in CoLA's auto-encoders and offer an effective-rank–aware recovery bound. We first explain the benefit of CoLA over standard low-rank approximations on linear projection layers. Then we (partially) extend the analysis to MLP layers. 

Let $n$ be the number of tokens, and $\sigma$ be a nonlinear activation function. Consider 
\begin{equation}\label{eq: E rho}
\Esigma(r):=\min_{\mat{A}\in\bR^{r\times\din},\mat{B}\in\bR^{\dout\times r}}\normF{\mat{Y-B\sigma(AX)}}.
\end{equation}
Here $\mat{X}\in\bR^{\din\times n}$ denotes the input to the linear layer in the compressed network, or equivalently the output of the already–compressed preceding layers. The target $\mat{Y}\in\bR^{\dout\times n}$ denotes the original output at the same layer, typically \(\mat{Y=W X_{\mathrm{True}}}\), where \(\mat{X_{\mathrm{True}}}\) is the input to this layer of the originally uncompressed model. In general, \(\mat{X}\) and \(\mat{X_{\mathrm{True}}}\) need not coincide; they may differ through a (possibly nonlinear) transformation induced by the preceding layers and their compression. If the activation $\sigma$ is the identity, problem~\eqref{eq: E rho} reduces to the conventional low-rank method:
\begin{equation*}
\Eone(r):=\min_{\mat{A}\in\bR^{r\times\din},\mat{B}\in\bR^{\dout\times r}}\mat{\normF{Y-BAX}}.
\end{equation*}
The following proposition shows that the optimal value with the nonlinear activation is no larger than in the identity case. Full proofs are in Appendix \ref{section: proof of main results}. 
\begin{proposition}\label{prop: erho <= eone}
    If $\sigma(0)=0$ and $\sigma'(0)\neq0$, then $\Esigma(r)\leq \Eone(r)$. 
\end{proposition}
Let $\mat{\row(X)}$ denote the row space of $\mat{X}$. Under the identity activation, we notice that the approximation is confined to $\mat{\row(X)}$.
The next result shows that, with a nonlinear activation, one can generate features $\mat{\sigma(u^\top X)}$ lying \emph{outside} $\mat{\row(X)}$; hence, CoLA can represent outputs that are not realizable by standard low-rank approximation.
\begin{proposition}\label{prop: rho Xu not in col(X)}
    Suppose that $\mat{X}\in\mathbb{R}^{\din\times n}$ has no identical columns, no zero columns and satisfies $n>\rank(\mat{X})$. If $\sigma(0)=0$, $\sigma'(0)\neq0$ and $\sigma''(0)\neq0$, then there exists $\mat{u}\in\mathbb{R}^{\din }$ such that $\sigma(\mat{u^\top X})$ is a nonzero vector and $\mat{\sigma(u^\top X)\notin \row(X)}$. 
\end{proposition}

Next we identify a sufficient data-dependent condition under which the CoLA layer  $\mat{B}\sigma(\mat{AX})$ strictly outperforms a standard low-rank layer $\mat{BAX}$, i.e., $\Esigma(r)<\Eone(r)$. Informally, if rows of $\mat{Y}$ lie substantially outside $\mat{\row(X)}$ and align with a nonlinear feature $\mat{v^\top:=\sigma(u^\top X)\notin\row(X)}$, then CoLA will achieve a strictly better approximation to $\mat{Y}$. To ground our discussion, we introduce the following notations. Let $P_\mat{X}$ denote the orthogonal projector onto $\mat{\row(X)}$ and set $P_{\mat{X}^\perp}:=I-P_{\mat{X}}$, where $I$ is the identity operator. Define $\mat{Y}_{\parallel}:=P_\mat{X}\mat{Y}$ and $\mat{Y}_{\perp}:=P_{\mat{X}^\perp}\mat{Y}$ (projectors are applied row-wisely to matrices). Similarly, let $P_{\mat{v}}$ denote the orthogonal projector onto $\operatorname{span}\{\mat{v}^\top\}$, and define $P_{\mat{v}^\perp}:=I-P_{\mat{v}}$. For a matrix $\mat{Z}$ with singular values $s_1\ge s_2\ge\cdots$, where $s_j:=0$ for $j>\rank(\mat{Z})$, write $s_{>k}(\mat{Z}):=\bigl(\sum_{j>k}s_j^2\bigr)^{1/2}$.

\begin{theorem}\label{theorem: Erho < Eid}
    Suppose that matrix $\mat{X}\in\mathbb{R}^{\din\times n}$ has no identical columns, no zero columns and satisfies $n>\rank(\mat{X})$. Suppose that $\sigma(0)=0$, $\sigma'(0)\neq0$ and $\sigma''(0)\neq0$. Let $\mat{u}\in\mathbb{R}^{\din}$ and $\mat{v^\top :=\sigma(u^\top X)\notin\row(X)}$. If \begin{equation}\label{eq: assumption strictly smaller}
        \normF{P_{\mat{v}^\perp}(\mat{Y})}^2<\normF{\mat{Y}_\perp}^2+\parens{s_{>r}(\mat{Y}_\parallel)}^2, 
    \end{equation}
    then $\Esigma(r)<\Eone(r)$.
\end{theorem}
In the extreme case where each row of $\mat{Y}$ lies in $\mathrm{span}\{\mat{v}^\top\}$, assumption \eqref{eq: assumption strictly smaller} holds trivially and $\Esigma(r)=0<\Eone(r)$. We also note that \eqref{eq: assumption strictly smaller} is sufficient (not necessary); sharper variants are possible.

We next assume that $\mat{Y}$ inherently admits an approximate autoencoder representation of $\mat{X}$, up to noise. The following theorem provides a non-asymptotic bound on the representation error of the $\Esigma(r)$ minimizer relative to this latent ground truth.


\begin{theorem}\label{theorem: effective rank}
    Suppose that there exist $\mat{A}_{\mathrm{True}}\in\bR^{r\times\din},\mat{B}_{\mathrm{True}}\in\bR^{\dout\times r}$ such that 
    \begin{equation}\label{eq: assumption XW - rhoBAX - G}
    \normTwo{\mat{Y}-\mat{B}_{\mathrm{True}}\sigma(\mat{A}_{\mathrm{True}}\mat{X})-\mat{G}}\leq \epsilon,
    \end{equation}
    where $\mat{G}$ is a random matrix of i.i.d Gaussian entries with zero-mean and variance $v^2$. Suppose that the optimal value $\Esigma(r)$ is obtained at $(\mat{A}^*,\mat{B}^*)$. Then with probability at least $1-2\exp(-(n+\dout))$, it holds that
    \begin{align*}
        &\Delta:=\normF{\mat{B_{\mathrm{True}}\sigma(A_{\mathrm{True}}X)-B^*\sigma(A^*X)}}\\
        \leq\ &\sqrt{r+r_\alpha(\mat{Y})}\parens{Cv\sqrt{n+\dout}+\epsilon+s_{r_\alpha(\mat{Y})+1}}\\
        &\quad +s_{>r_\alpha(\mat{Y})}(\mat{Y})+\Esigma(r),
    \end{align*}
    where $\alpha\in(0,1]$, $s_{r_\alpha(\mat{Y})+1}$ is $(r_\alpha(\mat{Y})+1)$-th largest singular value of $\mat{Y}$ and $C$ is an absolute constant. 
\end{theorem}


We note that the recovery bound explicitly depends on the effective rank $r_\alpha(\mat{Y})$, which is often empirically small (see Section \ref{sec: effective rank}). In particular, setting $\alpha=1$ reduces our result to a full-rank bound as $r_1(\mat{Y})=\rank(\mat{Y})$. When $\mat{Y}$ has a concentrated spectrum (i.e., $r_\alpha(\mat{Y})\ll\rank(\mat{Y})$), Theorem~\ref{theorem: effective rank} generally yields a tighter bound than the full-rank case. In addition, the established error bound reflects the role of the nonlinear activation through the term $\Esigma(r)$. As shown in Theorem~\ref{theorem: Erho < Eid}, under suitable conditions, $\Esigma(r)$ can be strictly smaller than its identity counterpart $\Eone(r)$, thereby yielding a smaller overall error bound.

\vspace{2pt}
\noindent{\bf Partial Extension to MLP Layers.} As stated in Section~\ref{subsec:CoLA}, we did not see a significant difference in performance when nonlinear activation was added on top of the auto-encoder layer [Eq~\eqref{eq:main-fwd}]. An auto-encoder followed by a non-linear activation is equivalent to just replacing the linear projection inside an MLP layer with Eq~\eqref{eq:main-fwd}, therefore our above theoretical analysis still holds. We still need more theoretical understanding of the case without activation after Eq~\eqref{eq:main-fwd}, which will be a future work.

\subsection{Computing Efficiency}
\label{sec:compute-eff}

\begin{table}[t]
\centering
\footnotesize
\begin{tabular}{c|c}
\toprule
\textbf{Operation} & \textbf{FLOPs} \\
\midrule
Attention: Q, K, V & $6nd^2$ \\
\midrule
Attention: SDP & $4n^2d$ \\
\midrule
Attention: Project & $2nd^2$ \\
\midrule
Feed-forward & $6ndd_{\text{ff}}$ \\
\midrule
Total Forward & $8nd^2 + 4n^2d +  6ndd_{\text{ff}} $ \\
\midrule
Total Backward & $16nd^2 + 8n^2d +  12ndd_{\text{ff}} $ \\
\bottomrule
\end{tabular}
\caption{Breakdown compute of a single LLaMA decoder layer in full-rank training. Lower-order terms such as bias, layer norm, activation are omitted.}
\label{tab:compute-breakdown-llama}
\vspace{-5pt}
\end{table}

\begin{table}[t]
\centering
\footnotesize
\resizebox{\linewidth}{!}{%
\begin{tabular}{c|c}
\toprule
\textbf{Methods} & \textbf{FLOPs} \\
\midrule
Full-Rank & \( C_{\text{Full-Rank}} = 24nd^2 + 12n^2d + 18ndd_{\text{ff}} \) \\
\midrule
CoLA & \( C_{\text{CoLA}} = 48ndr + 12n^2d + 18nr(d + d_{\text{ff}})\) \\
\midrule
(Re)LoRA & \( C_{\text{LoRA}} = C_{\text{CoLA}} + 16nd^2 + 12n^2d + 12ndd_{\text{ff}}\) \\
\midrule
SLTrain & \( C_{\text{SLTrain}} = C_{\text{Full-Rank}} + 24d^2r + 18dd_{\text{ff}}r \) \\
\midrule
GaLore & \( C_{\text{GaLore}} = C_{\text{Full-Rank}} + 16d^2r + 12dd_{\text{ff}}r \) \\
\bottomrule
\end{tabular}%
}
\caption{Estimated computing cost of a single LLaMA decoder layer. Results combine forward, backward and any additional compute occurred at optimizer step.}
\label{tab:compute-all-methods}
\vspace{-10pt}
\end{table}

We analyze and compare the computational complexity of CoLA with other pre-training methods based on the LLaMA architecture. We adopt a similar notion from \cite{kaplan2020scaling}, where a general matrix multiply (GEMM) between an $M\times N$ matrix and an $N\times K$ matrix involves roughly $2MNK$ add-multiply operations. We denote the model inner width as $d$, and the inner width of the feed-forward layer as $d_{\text{ff}}$. For simplicity, we only show non-embedding calculations of a single sequence with token batch size of $n$ for each decoder layer. This is because the total computation scales only linearly with the number of layers $n_{\text{layer}}$ and the number of sequences $n_{\text{seq}}$. Furthermore, lower-order cheap operations of complexity $\mathcal{O}(nd)$ or $\mathcal{O}(nd_{\text{ff}})$ are omitted, such as bias, layer norm, non-linear function, residual connection, and element-wise product. 

We show the detailed cost of the full-rank training in Table.~\ref{tab:compute-breakdown-llama}. Notice that we apply the $2\times$ rule when calculating the backward cost. This is because for each forward GEMM that Eq.~\eqref{eq:full-rank-fwd} describes, two GEMMs are needed to compute gradients for both the weight matrix $\mat{W}$ and the input $\mat{x}$, and are of the same cost the forward GEMM, i.e.,
\begin{equation}
    \nabla_{\mat{x}} = \mat{W}^{T}\nabla_{\mat{h}}, \nabla_{\mat{W}} = \nabla_{\mat{h}}\mat{x}^T.
\end{equation}
We apply the same analysis to all the following pre-training methods: 
\begin{itemize}[leftmargin=*]
    \item {\bf LoRA/ReLoRA} \cite{hu2021lora, lialin2023relora}: \(\mat{h}_{\text{LoRA}} = \mat{W}_0\mat{x} + \mat{BAx}\), with fixed $\mat{W}_0$.
    \vspace{-5pt}
    \item {\bf SLTrain} \cite{han2024sltrain}: \(\mat{h}_{\text{SLTrain}} = \mat{BAx} + \mat{Sx} = (\mat{BA} \oplus_{\mathcal{I}} \mathcal{V})\mat{x}\), where $\oplus$ denotes the scatter-add operator, $\mathcal{I}$ and $\mathcal{V}$ are the indices and values of non-zero elements in the sparse matrix $\mat{S}$.
    \vspace{-5pt}
    \item {\bf GaLore} \cite{zhao2024galore}: \(\mat{R}_t = \mat{P}_t^T \mat{G}_t,\)  \(\tilde{\mat{G}}_t = \mat{PN}_t\), where $\mat{P}_t$ projects the gradient $\mat{G}_t$ onto a low-rank space, and then projects it back when updating the full-rank weight $\mat{W}$.
\end{itemize}

We summarize the computational costs of these methods in Table~\ref{tab:compute-all-methods} and observe that the costs of SLTrain and GaLore are lower bounded by full-rank training, while (Re)LoRA is lower bounded by CoLA when choosing the same rank. In contrast, CoLA reduces the computation from full-rank training when $r < 0.62d$, assuming $d_{\text{ff}} \approx2.5d$ in LLaMA-like architecture. The default rank choice is set to $r = \frac{1}{4}d$, leading to a reduction in compute to about half the full-rank training. We refer all details of compute analysis to Appendix~\ref{sec:appendix-compute-analysis}.

\section{CoLA-M: A Memory-Efficient Implementation}

\begin{figure}[t]
\vspace{-10pt}
  \includegraphics[width=\columnwidth]{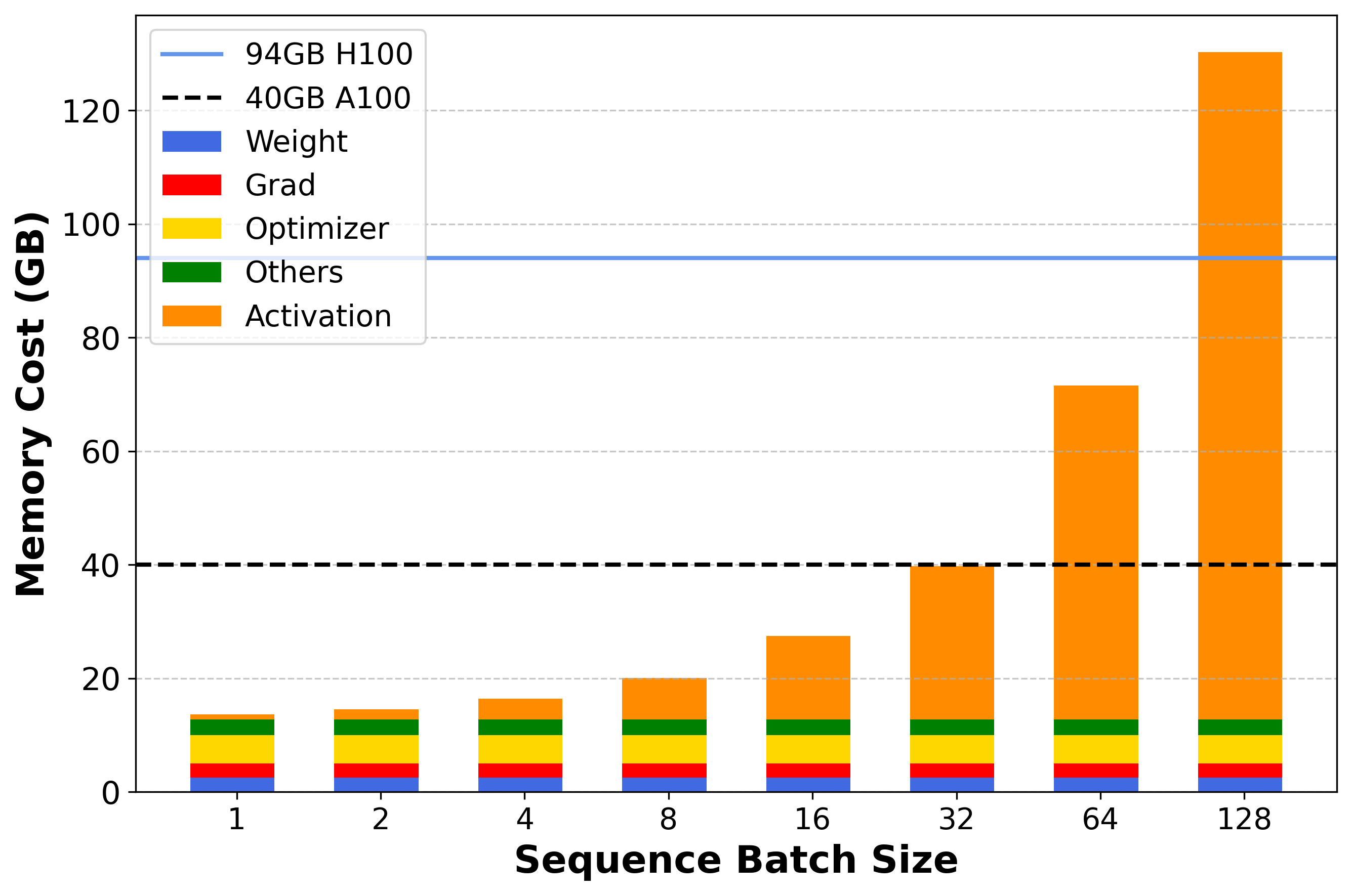}
  \caption{Memory breakdown for LLaMA-1B using fairly large sequence batch sizes in pre-training. The activation memory is at dominant place.}
  \label{fig:memory-break}
\end{figure}


In this section, we design and develop CoLA-M, a memory-efficient implementation to leverage CoLA's structural advantage to achieve superior memory saving without sacrificing throughput.


\subsection{Memory Breakdown in Pre-Training}

\begin{figure}[t]
  \includegraphics[width=\columnwidth]{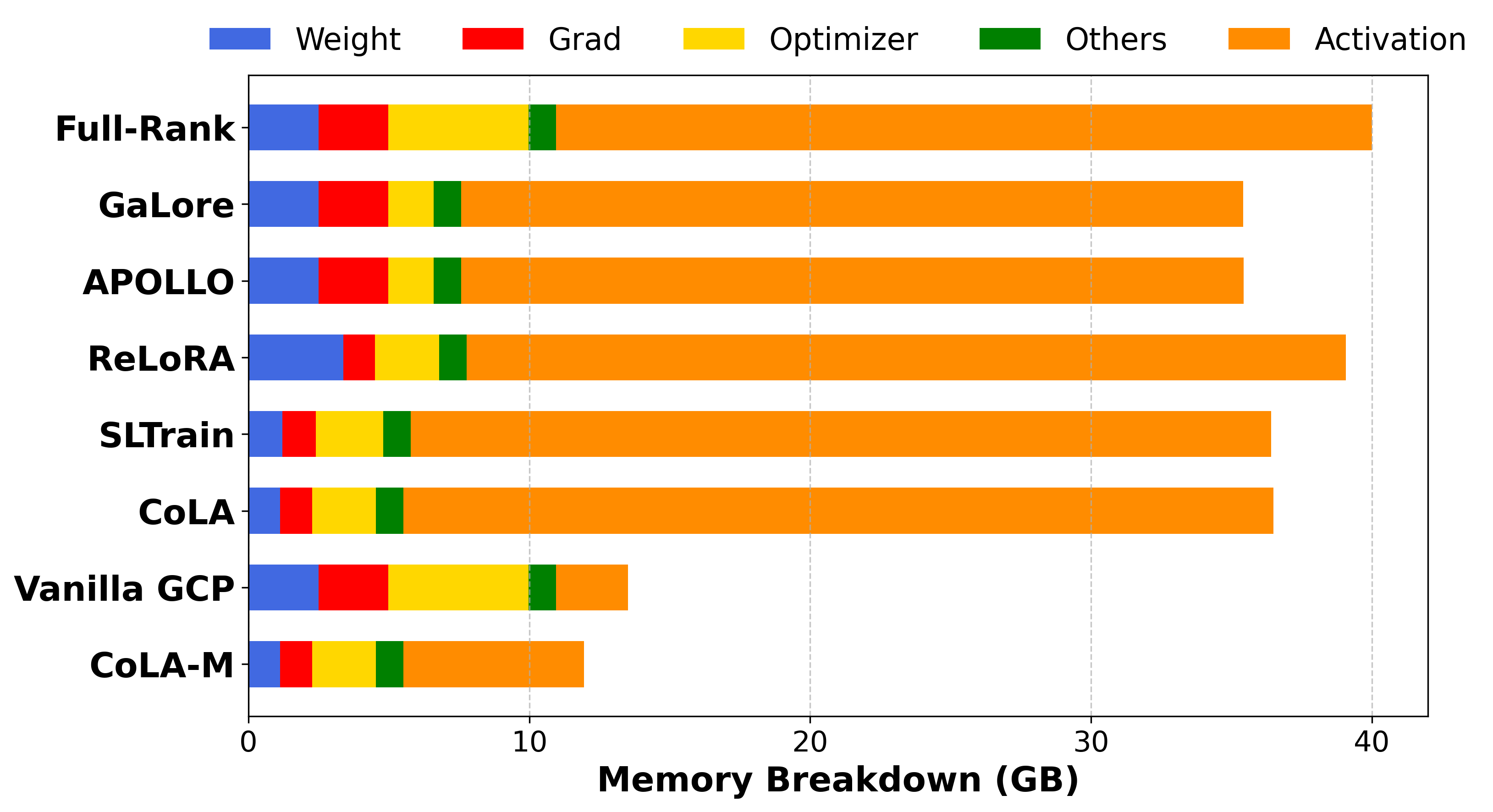}
  \caption{Memory breakdown of pre-training LLaMA-1B on single GPU using different pre-training methods.}
  \label{fig:gcp-memory-break}
  \vspace{-10pt}
\end{figure}

We assume a common notion that training modern transformers with Adam (or AdamW) involves four key memory components \cite{zhao2024galore, han2024sltrain}: model parameters (\(1\times\)), gradients (\(1\times\)), optimizer states (\(2\times\)), and activations (\(1\sim 4\times\)). We focus on the scenario where the memory cost determined by the model size is not on the extreme limit of the GPU. We argue that this is rather realistic, since the model size and the minimum required tokens should scale up simultaneously during pre-training \cite{kaplan2020scaling, hoffmann2022training, krajewski2024scaling, kumar2024scaling}. A tiny batch size on a single GPU would be impractical. Therefore, we analyze memory usage on a 40-GB A100 or a 94-GB H100 GPU with a fairly large sequence batch size. Fig.~\ref{fig:memory-break} \& \ref{fig:gcp-memory-break} show that activations dominate memory usage.



\subsection{CoLA Enables Efficient Checkpointing}
\label{sec:cola-m-efficiency}

\begin{figure}[t]
    \vspace{-15pt}
  \includegraphics[width=\columnwidth]{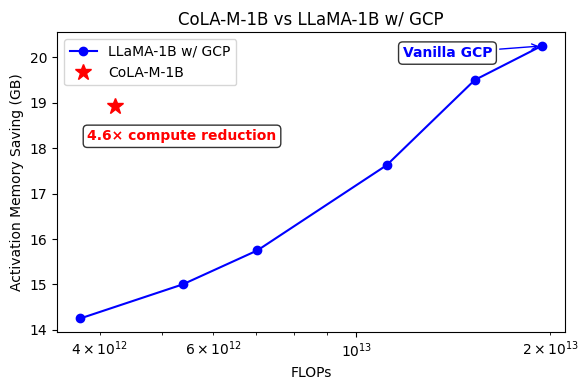}
  \caption{We show how memory reduction scales with the re-computation in full-rank training with GCP and compare with CoLA-M. With similar gains on memory efficiency, CoLA-M effectively reduces re-compute by $4.6\times$, enabling compute efficient checkpointing.}
  \label{fig:cola-m-vs-gcp}
  \vspace{-10pt}
\end{figure}

\begin{table*}[t]
\centering
\small
\vspace{-10pt}
\resizebox{\linewidth}{!}{%
\begin{tabular}{l|ccc|ccc|ccc|ccc}
\toprule
& \multicolumn{3}{c|}{60M} & \multicolumn{3}{c|}{130M} & \multicolumn{3}{c|}{350M} & \multicolumn{3}{c}{1B} \\
\midrule
\multicolumn{1}{l|}{\textit{r / d}} 
    & \multicolumn{3}{c|}{128 / 512}
    & \multicolumn{3}{c|}{256 / 768}
    & \multicolumn{3}{c|}{256 / 1024}
    & \multicolumn{3}{c}{512 / 2048} \\
\multicolumn{1}{l|}{\textit{Tokens}} 
    & \multicolumn{3}{c|}{1.1B}
    & \multicolumn{3}{c|}{2.2B}
    & \multicolumn{3}{c|}{6.4B}
    & \multicolumn{3}{c}{13.1B} \\
\midrule
& PPL & Param (M) & Mem (GB) & PPL & Param (M) & Mem (GB) & PPL & Param (M) & Mem (GB) & PPL & Param (M) & Mem (GB) \\
\midrule
Full-rank & 34.06 & 58 & 0.43 & \textbf{24.36} & 134 & 1.00 & \textbf{18.80} & 368 & 2.74 & 15.56 & 1339 & 9.98 \\
ReLoRA & 37.04 & 58 &  0.37 & 29.37 & 134 & 0.86 & 29.08 & 368 & 1.94 & 18.33 & 1339 & 6.79 \\
GaLore & 34.88 &  58 & 0.36 & 25.36 & 134 & 0.79 & 18.95 & 368 & 1.90 & 15.64 & 1339 & 6.60 \\
SLTrain & 34.15 & 44 & 0.32 & 26.04 & 97 & 0.72 & 19.42 & 194 & 1.45 & 16.14 & 646 & 4.81 \\
\midrule
CoLA  & {\bf 34.04} & {\bf 43} &  {\bf 0.32} & {24.48} & {\bf 94} &  {\bf 0.70}  & {19.40} & {\bf 185}  &  {\bf 1.38}  & {\bf 15.52} & {\bf 609}  & {\bf 4.54} \\
\bottomrule
\end{tabular}%
}
\caption{Comparison across various efficient pre-training methods of validation perplexity (PPL ($\downarrow$)), number of parameters in millions (Param), and the estimated memory usage (Mem) including model, gradient and optimizer states based on BF16 precision. We pre-train LLaMA models from 60M to 1B on the C4 dataset \cite{raffel2020exploring} following the same setup and compare results directly against those reported in \cite{zhao2024galore, han2024sltrain}.}
\label{tab:main-results}
\vspace{-10pt}
\end{table*}

Gradient checkpointing (GCP) \cite{chen2016training} is a system-level technique that reduces memory usage by selectively storing (“checkpointing”) only a subset of intermediate results during the forward pass. When the backward pass begins, the missing activations are recomputed on the fly instead of being stored in memory, thereby lowering the memory cost. A vanilla (also the most effective) implementation of GCP in LLM pre-training is to save merely the input and output of each transformer block, and re-compute everything within each block during the backward step. Some works have investigated the optimal selection of checkpoints through both empirical and compiler view \cite{feng2021optimal, he2023transcending}. Such techniques can also be developed for CoLA, and are beyond the scope of this paper.

Motivated by the bottleneck structure of CoLA, we implement CoLA-M as {\bf saving only the low-rank activations} (red circles in Fig.~\ref{fig:block}), and re-compute the up projections, and (if applicable) the self-attention (painted in sketch in Fig.~\ref{fig:block}) during the backward pass. This reduces the re-computation cost to half of the CoLA forward. We refer the detailed analysis to Appendix~\ref{sec:appendix-mem-analysis}.

Although delicate optimizations of GCP is beyond our scope, we show in Fig.~\ref{fig:cola-m-vs-gcp} the quantitative results and scaling behavior of GCP on LLaMA-1B when applying a heuristic checkpointing strategy. CoLA-M greatly reduces the re-computation cost by $\bf 4.6 \pmb{\times}$ while achieving similar memory saving (18.94GB) as vanilla GCP (20.25GB).

\section{Experiments}
\label{sec:results}

\begin{table}[t]
\centering
\small
\resizebox{\linewidth}{!}{%
\begin{tabular}{c|c|c|c|c|c|c}
\toprule
& Mem (GB) & 10k & 40k & 80k & 120k & 150k \\
\midrule
8-bit Adam & 72.59 & N/A & 18.09 & 15.47 & 14.83 & 14.61  \\
\midrule
8-bit GaLore & 65.16 & 26.87 & 17.94 & 15.39 & 14.95 & 14.65  \\
\midrule
SLTrain & 60.91 & 27.59 & \multicolumn{4}{c}{N/A} \\
\midrule
CoLA-M & {\bf 26.82} & {\bf 22.76} & {\bf 16.21} & {\bf 13.82} & {\bf 13.09} & {\bf 12.73} \\
\bottomrule
\end{tabular}%
}
\vspace{-5pt}
\caption{Validation perplexity of LLaMA-7B pre-trained on C4 dataset. 8-bit Adam/GaLore are collected from \cite{zhao2024galore}. SLTrain is collected from \cite{han2024sltrain}. No results of BF16 Adam reported.}
\label{tab:result-7b}
\vspace{-5pt}
\end{table}

\begin{table}[t]
\centering
\small
\resizebox{\linewidth}{!}{%
\begin{tabular}{c|cc|cc|cc}
\toprule
& \multicolumn{2}{c|}{60M} & \multicolumn{2}{c|}{130M} & 
\multicolumn{2}{c}{350M} \\
\cmidrule{2-7}
& PPL & FLOPs & PPL & FLOPs & PPL & FLOPs \\
\midrule
Full-Rank & 34.06 & $1\times$ & 24.36 & $1\times$ & 18.80 & $1\times$ \\
\midrule
Control & 37.73 & $0.4\times$ & 27.05 & $0.5\times$ & 20.53 & $ 0.4\times$  \\
\midrule
\multirow{2}{*}{CoLA} & 34.04 & $0.4\times$ & 24.48 & $0.5\times$ & 19.40 & $0.4\times$ \\
& {\bf 31.52} & $0.7\times$ & {\bf 23.97} & $0.7\times$ & {\bf 18.32} & $0.7\times$ \\
\bottomrule
\end{tabular}%
}
\vspace{-5pt}
\caption{Scaling behavior of CoLA and full-rank training. Control represents scaling down the full-rank training cost to be similar with CoLA in default, by reducing number of layers and/or size down model width.}
\label{tab:cola-scaling-up}
\vspace{-10pt}
\end{table}

\begin{table*}[t]
\vspace{-10pt}
\centering
\small
\begin{tabular}{c|c|c|c|c|c|c|c|c|c|c}
\toprule
& Pre-Training Loss & QQP & SST-2 & MRPC & COLA & QNLI & MNLI & RTE & STS-B & GLUE Avg \\
\midrule
BERT$_{\text{Large}}$ & 1.263 & 91.1 & 92.1 & {\bf 90.7} & 53.1 & 91.6 & 84.3 & 69.9 & 88.9 & 82.7\\
\midrule
CoLA & {\bf 1.257} & {\bf 91.2} & {\bf 92.3} & 90.6 & {\bf 54.1} & {\bf 91.7} & {\bf 84.3} & {\bf 74.2} & {\bf 89.7} & {\bf 83.5} \\
\bottomrule
\end{tabular}%
\caption{Fine-tuning CoLA and BERT$_{\text{Large}}$ on GLUE. Both models are fine-tuned for three epochs. F1 scores are reported for MRPC, Pearson correlations are reported for STS-B, Matthews correlations are reported for COLA (task), accuracies are reported for all other tasks. Reported metrics are the mean of 5 best out of 10 random seeds.}
\label{tab:bert-ft}
\vspace{-5pt}
\end{table*}

\subsection{Pre-Training within Compute-Optimal}
\label{llama-pt}
We validate our proposed methods by extensively pre-training LLaMA-like LLMs from 60M to 7B scales following {\bf the exact experimental setup} in \cite{zhao2024galore, han2024sltrain}. Trainings were done using C4 dataset \cite{raffel2020exploring} without data repetition on roughly compute-optimal\footnote{Compute optimal regime refers to the token-to-parameter (T2P) ratio being \textasciitilde 20~\cite{hoffmann2022training}.} amounts of tokens. We compare CoLA with baselines including {\bf full-rank} pre-training, {\bf ReLoRA} \cite{hu2021lora}, \textbf{GaLore} \cite{zhao2024galore}, and \textbf{SLTrain} \cite{han2024sltrain}, with a focus on methods that explore model efficiency. 

We implement CoLA and CoLA-M by parameterizing all MLP layers and all projection layers in attention with  auto-encoders [i.e. Eq.~\eqref{eq:main-fwd}], and keep all other parameters and operations unchanged. We use AdamW optimizer and cosine annealing learning rate scheduler \cite{loshchilov2016sgdr} with warm-up. We show details to Appendix~\ref{sec:appendix-llama-pt}.

Table~\ref{tab:main-results} compares our methods and other efficient pre-training techniques in terms of validation perplexity, parameter size, and estimated memory usage of model, gradients and optimizer states. CoLA has the {\bf smallest model size}, thereby {\bf consumes the least memory}, and {\bf performs on-par with full-rank} baselines. CoLA {\bf uniformly surpasses} other efficient training baselines in both {\bf efficiency} and {\bf accuracy}. Table~\ref{tab:result-7b} compares the validation perplexity on the 7B model for 150k steps\footnote{Due to resources constraints, 7B models are trained below compute optimal budget~\cite{zhao2024galore,han2024sltrain}.}. CoLA(-M) significantly outperforms 8-bit Adam/GaLore by {\bf 12.73} vs \textasciitilde 14.6, while saving two-third memory.

\vspace{2pt}
\noindent{\bf Scaling Behavior: }Table~\ref{tab:cola-scaling-up} shows how CoLA might be improved when compute is scaled up. The default rank choices reduce half the computing cost, without harming the model performance. Meanwhile, if we relax the computing restriction and moderately increase the rank, then CoLA outperforms full-rank training in all three scales, while still being fairly smaller and reducing the computing cost. One might argue that full-rank training can also be scaled down to a similar computing cost of CoLA and might perform similarly. We implement such baselines in Table~\ref{tab:cola-scaling-up} and refer this setup to ``Control". We typically reduce the number of layers or the model width of full-rank models to scale down their computing cost. We find empirically that they increase perplexity (PPL) significantly and dramatically underperform CoLA.

\subsection{Pre-Training beyond Compute-Optimal}

According to Chinchilla scaling law~\cite{hoffmann2022training}, compute-optimal training is at the efficient frontier when given a fixed computing budget or a target model size. However, leading industrial groups with massive computing resources tend to extensively overtrain smaller models for efficient deployment, such as LLaMA-3~\cite{grattafiori2024llama} 1-3B models being trained up to 9 Trillion tokens. To evaluate CoLA's effectiveness beyond the compute-optimal regime, we further experiment the following two over-training settings.

\vspace{2pt}
\noindent{\bf LLaMA-350M with 51B Tokens: }
We prolong the training duration by 8$\times$ of the compute-optimal budget for both CoLA\footnote{\label{note:0.7-cola}We choose CoLA at $0.7\times$ compute of full-rank baseline, as its superior performance observed in Table~\ref{tab:cola-scaling-up}.} and full-rank LLaMA at 350M scale. This results in 51B total training tokens. CoLA continues outperforming full-rank baseline on validation perplexity of {\bf 13.96} vs 14.47, consistent with results at compute-optimal observed from Table~\ref{tab:cola-scaling-up}.

\vspace{2pt}
\noindent{\bf BERT$_{\text{Large}}$ (350M) with 85B Tokens:} We adopt the exact infrastructure and training configurations from NVIDIA's faithful BERT~\cite{devlin2019bert} reproduction\footnote{\label{nv-rep}See details at \href{https://github.com/NVIDIA/DeepLearningExamples}{NVIDIA's official Github repo}.} and pre-train both CoLA\footref{note:0.7-cola}\textsuperscript{,}\footnote{See detailed configurations in Appendix~\ref{sec:appendix-bert-pt}} and full-rank BERT$_{\text{Large}}$ at 350M scale on Wikipedia for 85B tokens. CoLA outperforms BERT$_{\text{Large}}$ on training loss of {\bf 1.257} vs 1.263. We fine-tune both pre-trained models for three epochs following~\cite{devlin2019bert} on GLUE~\cite{wang2018glue} benchmark and show results in Table~\ref{tab:bert-ft}. CoLA outperforms full-rank baseline across 7 out of 8 tasks, and on average score of {\bf 83.5} vs 82.7.

These results further demonstrate CoLA's effectiveness across both {\bf encoder/decoder} architectures, both {\bf compute-optimal/over-train} settings, and different activations (GeLU and Swish).

\subsection{Training/Inference System Performance}

\begin{figure}[t]
    \vspace{-5pt}
  \includegraphics[width=\columnwidth]{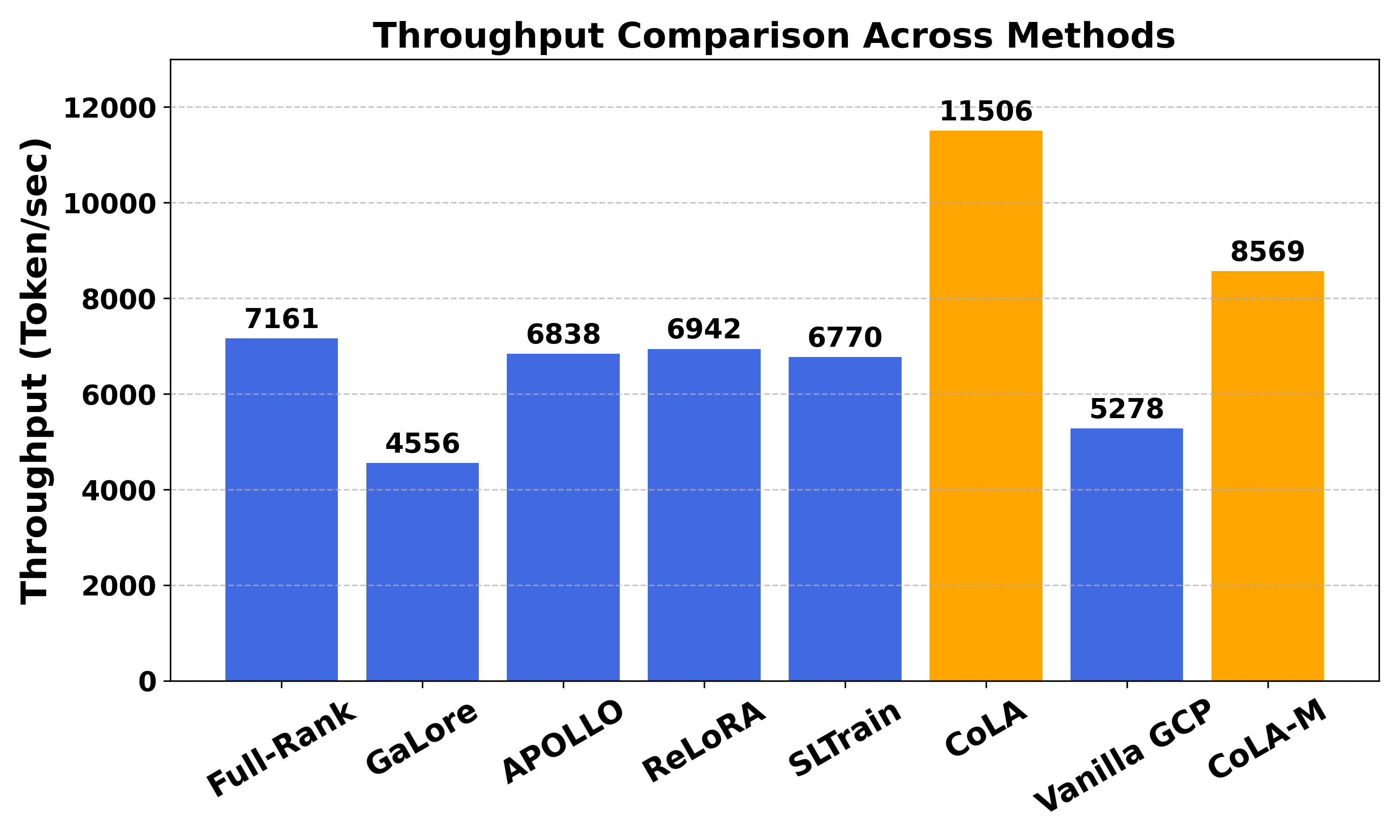}
  \caption{Comparison of throughput measured when pre-training a LLaMA-1B on a 40 GB A100 GPU with sequence batch size of 16 for different methods. }
  \label{fig:throughput}
  \vspace{-10pt}
\end{figure}

\begin{table}[t]
\centering
\small
\resizebox{\linewidth}{!}{%
\begin{tabular}{c|ccc|ccc}
\toprule
& \multicolumn{3}{c|}{1B (BZ = 64)} & 
\multicolumn{3}{c}{7B (BZ = 16)} \\
\cmidrule{2-7}
& Mem (GB) & Token/s & FLOPs & Mem (GB) & Token/s & FLOPs\\
\midrule
Full-Rank & 69.84 & 12,365 & $1\times$ & 84.94 & 5,810 & $1\times$ \\
\midrule
Vanilla GCP & {\bf 14.89} & 8,799 & $1.68\times$ & 52.49 & 4,357 & $1.67\times$ \\
\midrule 
CoLA & 66.46 & {\bf 22,979} & $\bf 0.40\pmb{\times}$ & 55.52 & {\bf 9,638} & $\bf 0.40\pmb{\times}$ \\
\midrule
{CoLA-M} & 17.33 & 16,617 & $0.55\times$ & {\bf 26.82} & 7,026 &$0.54\times$ \\
\bottomrule
\end{tabular}
}
\caption{Detailed measurements and comparison of CoLA and CoLA-M against full-rank and vanilla GCP on a 94 GB H100 GPU. CoLA-M consumes only one third of the memory while achieving higher throughput than full-rank training with only about half its compute.}
\label{tab:cola-gcp-mem-flops}
\vspace{-10pt}
\end{table}

\paragraph{Superior Training Efficiency.} We further validate CoLA’s efficiency from a practical perspective: CoLA delivers superior out-of-the-box system performance compared to full-rank and other efficient training methods. Fig. \ref{fig:throughput} compares pre-training throughput for the 1B-scale LLaMA model (batch size 16, fully utilizing A100 GPUs). Among evaluated methods, only CoLA and CoLA-M surpass the full-rank baseline throughput. Notably, CoLA-M maintains higher throughput despite recomputation overhead, significantly outperforming vanilla GCP. Table~\ref{tab:cola-gcp-mem-flops} provides detailed measurements, showing CoLA-M cuts computing cost nearly by half and reduces memory usage by two-thirds, achieving great balance between memory and compute efficiency. Profiling details are available in Appendix~\ref{sec:appendix-detailed-experiment-setting}.
\vspace{-1ex}
\paragraph{Superior Inference Efficiency.} CoLA also speeds up inference and reduces memory cost. Table~\ref{tab:inference} (Appendix~\ref{sec:inference}) shows that CoLA off-the-shelf reduces inference latency and memory cost by up to $\bf 1.64\pmb{\times}$ and $\bf 1.67\pmb{\times}$, respectively.




\section{Conclusions}
We have proposed CoLA, and its memory efficient variant CoLA-M, to achieve collectively {\bf parameter}, {\bf computing} and {\bf memory efficiency} in both pre-training and inference for large foundation models. CoLA has reduced $\bf 2\pmb{\times}$ model size and computing cost while preserving full-rank level performance. CoLA-M trades minimum overhead for state-of-the-art memory reduction, while still improving training throughput over full-rank baselines. CoLA is promising to save substantial GPU resources in LLM industry. This work has focused on dense architectures. In the future, it is worth extending CoLA to the mixture-of-expert (MoE) architecture.


\section{Limitations}
Most of our pre-training experiments follow the exact setup in \cite{zhao2024galore, han2024sltrain} and are conducted in the widely accepted computing-optimal setting~\cite{hoffmann2022training} under academic budget. Therefore, they are not trained with the same amount of tokens as industry-produced models. However, our BERT$_{\text{Large}}$ experiment follows NVIDIA's faithful reproduction and is directly compared with the reproduced BERT$_{\text{Large}}$ on standard downstream tasks (e.g., GLUE). CoLA outperforms BERT$_{\text{Large}}$ and shows great potential for producing competitive models. We have also pre-trained the LLaMA-350M with a high token-to-parameter ratio, showing that CoLA consistently outperform full-rank pre-training in terms of both accuracy and efficiency. 


\section*{Acknowledgments}

This material is based upon work supported by
the U.S.\ Department of Energy, Office of Science, Office of Advanced
Scientific Computing Research, Artificial Intelligence for Science program, under contracts DE-SC0025390 and DE-AC02-06CH11357.

This research used resources of the National Energy Research Scientific Computing Center, a DOE Office of Science User Facility
supported by the Office of Science of the U.S. Department of Energy under Contract No. DE-AC02-05CH11231 using NERSC award ASCR-ERCAP0030039, as well as NERSC award ALCC-ERCAP0031379.

\bibliography{main}

\appendix

\section{Observation of Low-Rank Activation in Pre-Trained GPT2}
\begin{figure}[h]
    \centering
    \includegraphics[width=\linewidth]{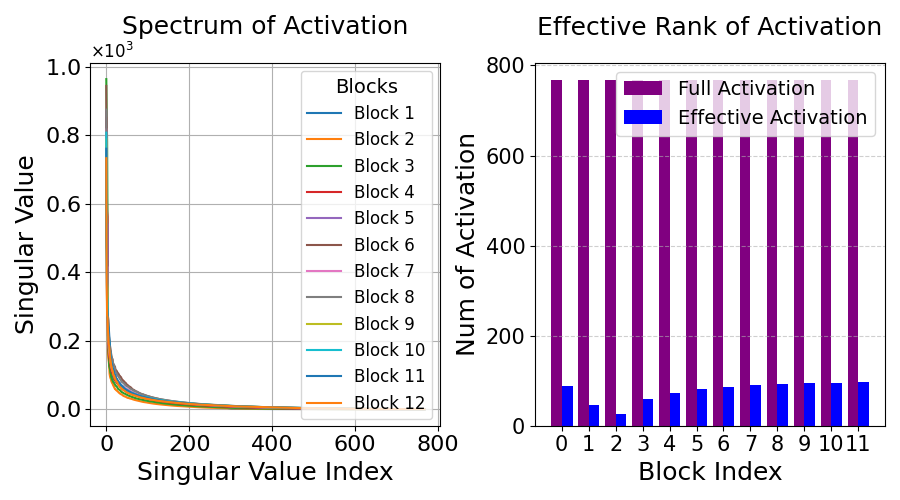}
    \caption{Spectrum of attention layer (query) output (i.e., $\mat{W_q x}$).}
    \label{fig:Q_Spectrum}
\end{figure}

\begin{figure}[h]
    \centering
    \includegraphics[width=\linewidth]{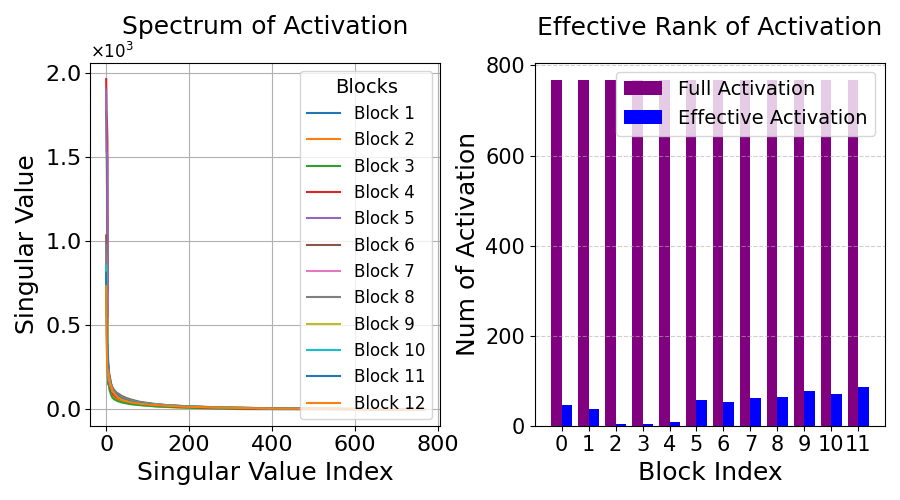}
    \caption{Spectrum of attention layer (key) output (i.e., $\mat{W_k x}$).}
    \label{fig:K_Spectrum}
\end{figure}

In this section, we further show the low-rank structure in model activations evaluated on a pre-trained GPT-2 \cite{radford2019language} small. The evaluation is conducted WikiText2 dataset with sequence length 1024. We fix $\alpha = 0.95$ throughout this section. Similar patterns are observed from the attention layers (Fig.~\ref{fig:Q_Spectrum},~\ref{fig:K_Spectrum},~\ref{fig:V_Spectrum}). The low-rank nature of activations is evident across all the different components of the model. This suggests that despite the high-dimensional representations, the effective dimensionality of the activations remains constrained.
\begin{figure}[h]
    \centering
    \includegraphics[width=\linewidth]{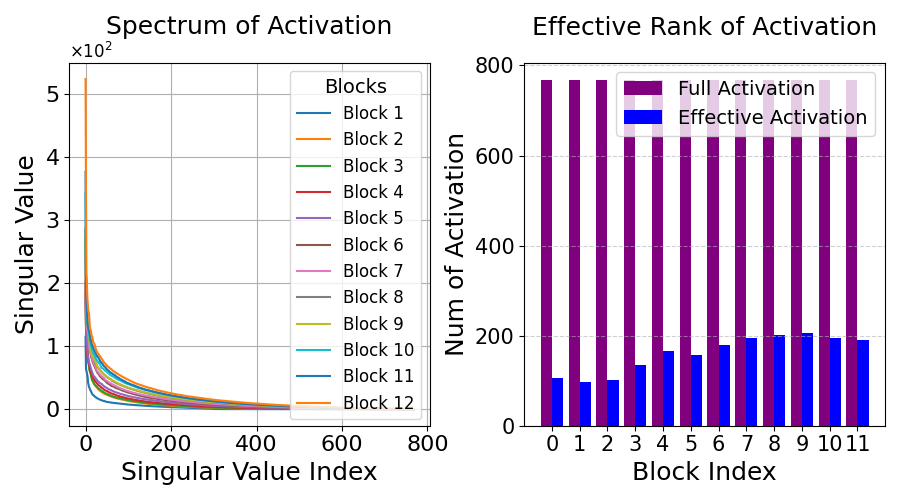}
    \caption{Spectrum of attention layer (value) output (i.e., $\mat{W_v x}$).}
    \label{fig:V_Spectrum}
\end{figure}
\begin{figure}[h]
    \centering
    \includegraphics[width=\linewidth]{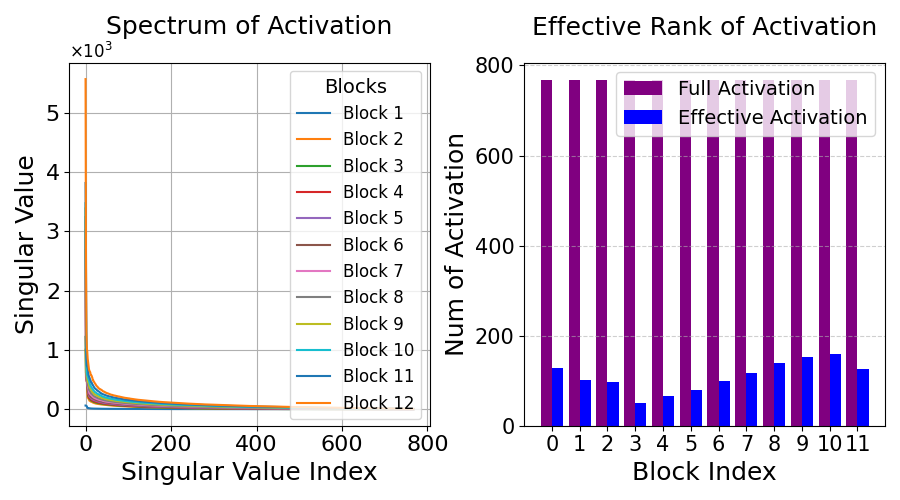}
    \caption{Spectrum of MLP block output (i.e., $\mat{W_2 \sigma (W_1 x)}$).}
    \label{fig:V_Spectrum}
\end{figure}

\label{sec:appendix-low-rank-activation}

\section{Detailed Compute Analysis}
\label{sec:appendix-compute-analysis}
According to Table.~\ref{tab:compute-breakdown-llama}, the total compute of full-rank training is simply combining forward and backward as 
\begin{equation}
    C_{\text{Full-Rank}} = 24nd^2 + 12n^2d + 18ndd_{\text{ff}}.
    \label{eq:full-rank-compute}
\end{equation}

In our proposed architecture, every single linear layer is replaced by low rank matrices $\mat{A}$,  $\mat{B}$, and an activation function sandwiched in between. The activation only introduces trivial compute thus can be omitted in the calculation. For each $d^2$ and $dd_{\text{ff}}$ in Eq.~\eqref{eq:full-rank-compute}, CoLA effectively converts them into $2dr$ and $r(d+d_{\text{ff}})$. Therefore the total compute of CoLA is
\begin{equation}
C_{\text{CoLA}} = 48ndr + 12n^2d + 18nr(d + d_{\text{ff}})
    \label{eq:cola-compute}.
\end{equation}
Plugging in an actual setting of LLaMA/CoLA-1B, in which $r = \frac{1}{4}d$ and $r \approx \frac{1}{10}d_{\text{ff}}$, we achieve a compute reduction from Eq.~\eqref{eq:full-rank-compute} to approximately
\begin{equation}
C_{\text{CoLA-1B}} = 16.5nd^2 + 12n^2d + 1.8ndd_{\text{ff}}.
\end{equation}

We now discuss and compare CoLA with other efficient pre-training methods in terms of their compute complexity. We start with LoRA \cite{hu2021lora} and ReLoRA \cite{lialin2023relora}. They share the same architecture that's shown in Fig.~\ref{fig:arc} a), in which low rank matrices $\mat{A}\in \mathbb{R}^{r \times d_{\text{in}}} $ and $\mat{B}\in \mathbb{R}^{d_{\text{out}} \times r} $ are adapted onto a full rank matrix $\mat{W}_0 \in \mathbb{R}^{d_{\text{out}}\times d_{\text{in}}}$. Hence modifies Eq.~\eqref{eq:full-rank-fwd} into
\begin{equation}
    \mat{h} = \mat{W}_0\mat{x} + \mat{BAx}.
    \label{eq:lora-fwd}
\end{equation}
This yields a consistently more expensive forward step than the full-rank training regardless the choice of $r$. During the backward step, since gradient does not flow into $\mat{W}_0$, only one GEMM that computes gradient w.r.t $\mat{x}$ is involved with the full-rank component $\mat{W}_0\mat{x}$. Combining together both full-rank and low-rank components in both forward and backward step, the total compute of LoRA is
\begin{multline}
    C_{\text{LoRA}} = 16nd^2 + 12n^2d + 12ndd_{\text{ff}} \\ + \underbrace{48ndr + 18nr(d + d_{\text{ff}})}_{C_{\text{CoLA}}}.
    \label{eq:lora-compute}
\end{multline}
When choosing the same $r$ for LoRA and CoLA, we have $C_{\text{LoRA}} > C_{\text{CoLA}} $ always true.

In ReLoRA \cite{lialin2023relora}, the hybrid strategy that warms up with the full-rank training arises more uncertainties in analyzing its complexity. And such strategy needs delicate tuning of hyper-parameters such as the full rank warm-up ratio, the restart frequency of optimizer, etc, and the choice of rank might also be affected by these strategy-level hyper-parameters. Therefore, we follow the same notion in \cite{zhao2024galore} that only consider the pure low-rank training of ReLoRA, which simplifies the compute analysis of ReLoRA to be the same as LoRA.

SLTrain \cite{han2024sltrain} proposes a low-rank + sparse parameterization  instead of having a fixed full-rank matrix $\mat{W}_0$. The architecture of SLTrain is shown in Fig.~\ref{fig:arc} c). We continue using the notation for the low-rank matrices, and denote the sparse matrix as $\mat{S}$, with the sparsity level as $\delta$. This modifies Eq.~\eqref{eq:full-rank-fwd} into
\begin{equation}
    \mat{h} = \mat{BAx} + \mat{Sx} = (\mat{BA} \oplus_{\mathcal{I}} \mathcal{V})\mat{x},
    \label{eq:sltrain-fwd}
\end{equation}
where $\oplus$ denotes the scatter-add operator, $\mathcal{I}$ and $\mathcal{V}$ denote the indices and values of non-zero elements in $\mat{S}$. This implementation avoids instantiating a full sized $\mat{S}$, instead keeping only the non-zero elements. However, this introduces non-trivial reconstruction cost of $\mat{BA}$ in every step. And if we further denote $\tilde{\mat{W}} = \mat{BA} \oplus_{\mathcal{I}} \mathcal{V}$, then the forward data-flow that starts from $\tilde{\mat{W}}$ is the same as in the full-rank training, as well as the backward data-flow that ends at $\tilde{\mat{W}}$. Therefore, the total compute of SLTrain should be $C_{\text{full-rank}}$ plus reconstructing $\tilde{\mat{W}}$, and its corresponding $2\times$ compute during backward, i.e.,
\begin{equation}
C_{\text{SLTrain}} = C_{\text{full-rank}} + 24d^2r + 18dd_{\text{ff}}r.
\end{equation}

For the last class of method to discuss, GaLore \cite{zhao2024galore} and it's follow-ups such as Fira \cite{chen2024fira} and APOLLO \cite{zhu2024apollo}, all investigate the memory efficiency associated with the AdamW optimizer. We only show the data-flow GaLore in Fig.~\ref{fig:arc} b), others are similar except some minor differences in how to manipulate gradients. The model architecture is kept unchanged in all these methods. Therefore, the complexity analysis is on the additional compute for projecting gradients into a low-rank space. GaLore proposes the following update rules:
\begin{equation}
\begin{aligned}
     \mat{R}_t &= \mat{P}_t^T \mat{G}_t, \tilde{\mat{G}}_t = \alpha\cdot \mat{PN}_t, \\
    \mat{W}_t &= \mat{W}_{t-1} + \eta \cdot \tilde{\mat{G}}_t,
\end{aligned}
\end{equation}
where the projector $\mat{P}_t\in \mathbb{R}^{d\times r}$ at time $t$ is computed by decomposing $\mat{G}_t\in \mathbb{R}^{d\times d}$ via singular value decomposition (SVD) and is updated periodically, $\mat{N}_t \in \mathbb{R}^{d\times r}$ is the low-rank optimizer states, $\alpha$ is a scaling factor and $\eta$ is the learning rate. Therefore, the total compute of GaLore is
\begin{equation}
    C_{\text{GaLore}} = C_{\text{full-rank}} + 16d^2r + 12dd_{\text{ff}}r.
\end{equation}

We remark that the compute analysis for the additional cost of SLTrain and GaLore (and its variants) is of limited scope and does not necessarily reflect their actual overhead. The actual cost will be dependent on other practical considerations on both algorithm and system level, such as the specific use case of these methods (e.g., pre-training, fine-tuning, etc), the actual number of the optimizer steps performed, the actual number of forward and backward steps performed when fixing total training tokens (i.e., if the hardware can afford larger batch sizes then the actual steps are fewer). It is almost impossible to give a unified notion while being fair when comparing between them. Hence we follow the similar setup used in \cite{zhao2024galore, han2024sltrain, chen2024fira, zhu2024apollo} when they analyze memory efficiency and measure system-level performance. However, it is rather safe to conclude that the overall cost introduced by GaLore and its variants will be diluted in real practices of pre-training due to the optimizer step is not frequent as forward and backward steps, hence are less expensive than SLTrain. Nonetheless, we highlight the fact that all the aforementioned methods are non-trivially more expensive than CoLA in terms of compute, and are all (except LoRA/ReLoRA) lower bounded by the full-rank training.

\section{Detailed Memory Analysis}
\label{sec:appendix-mem-analysis}

\begin{table}[t]
\centering
\small
\resizebox{\linewidth}{!}{%
\begin{tabular}{c|c|c}
\toprule
\textbf{Methods} & \textbf{Memory} & \textbf{Re-Compute} \\
\midrule
Full-Rank & \( 20nd + 2n^2h \) & N/A \\
\midrule
Vanilla GCP & \( nd \) & \( 23nd^2 + 4n^2d \) \\
\midrule
CoLA & \( 17.5nd + 2n^2h + 14nr \) & N/A \\
\midrule
CoLA-M & \( 2nd + 7nr \) & \( 18.5ndr + 4n^2d \) \\
\bottomrule
\end{tabular}%
}
\caption{Memory and re-computation analysis of full-rank training with vanilla GCP vs. CoLA and CoLA-M.}
\label{tab:compute-memory-gcp}
\end{table}

We analyze the memory and re-computation cost using the same notions as in Section~\ref{sec:compute-eff} and denote $h$ as the number of attention heads. We further simplify the analysis under LLaMA architecture by uniformly assuming $d_{\text{ff}} \approx 2.5d$. We start with the activation memory of full-rank training:
\begin{multline}
M_{\text{full-rank}} = \underbrace{3nd}_{\mat{Q}, \mat{K}, \mat{V}} + \underbrace{2n^2h + 2nd}_{\text{attention}} + \underbrace{11nd}_{\text{ffw}} \\
\underbrace{2nd}_{\text{residual connection}} + \underbrace{2nd}_{\text{layer norm}} = 20nd + 2n^2h.
\label{eq:full-rank-act-mem}
\end{multline}
When applying vanilla GCP, only the output of each block is saved, and all other activations are re-computed when needed. This dramatically reduces the total activation memory to only
\begin{equation}
    M_{\text{vanilla-GCP}} = nd.
\end{equation}
However, such benefit comes with a cost equal to almost an entire forward step. From Table. \ref{tab:compute-breakdown-llama}, we have the cost of vanilla-GCP as
\begin{equation}
    C_{\text{vanilla-GCP}} = C_{\text{full-rank}} +  23nd^2 + 4n^2d.
\end{equation}
Although we mentioned that delicate optimization of vanilla-GCP is beyond the scope of our discussion, we show a heuristic strategy when selecting checkpoints. Refer to Eq.~\eqref{eq:full-rank-act-mem}, activations that associated with minimal re-compute are: layer norm, residual connection, and non-linear function (included in the ffw term). Then intuitively these activations should always be re-computed when trying to save memory. In fact this can save a fair amount of memory. Note in this paper we analyze compute in pure theoretical notion that lower order terms does not bring noticeable effect hence are omitted. In practice, however, re-computation brings latency even for theoretically trivial operations, and will lower the overall GPU throughput. Other terms in Eq.~\eqref{eq:full-rank-act-mem} are all significant components when mapping to FLOPs change. One can gradually add more operations into the re-compute list and trade for more memory savings. We show the trend how they scale in Fig.~\ref{fig:cola-m-vs-gcp}.

Now we discuss CoLA and how it enables compute efficient checkpointing. We first evaluate how much memory overhead introduced by the low-rank activations. Compared to Eq.~\eqref{eq:full-rank-act-mem}, CoLA adds $2nr$ for each of the low-rank layers, i.e., $nr$ for $\mat{Ax}$, another $nr$ for $\sigma(\mat{Ax})$, thereby 
\begin{multline}
M_{\text{CoLA}} = M_{\text{full-rank}} + \underbrace{14nr}_{\text{low-rank }\sigma} - \underbrace{2.5nd}_{\text{remove original }\sigma}
\end{multline}
We notice that when model scales up, the original LLaMA activation no longer brings benefit to model performance, hence can be removed, which corresponds to $2.5nd$ less activations.

As shown in Figure. \ref{fig:block}, 
CoLA has multiple non-linear functions injected along the normal data-flow. This partitions the previously longer path, i.e., the whole block, to significantly shorter paths bounded by these low-rank activations. This provides a natural selection of checkpoints that are of $r$-dimensional instead of $d$. More importantly, these shorter paths halve the re-compute steps. We show in Figure. \ref{fig:block} that only the weights that are painted in sketch need re-computation during the backward step of CoLA-M. This reduces significantly the cost of implementing GCP in CoLA-like architecture, results in the cost of only
\begin{equation}
    C_{\text{CoLA-M}} = C_{\text{CoLA}} + 18.5ndr + 4n^2d.
\end{equation}
Meanwhile, the memory saving of CoLA-M is still significant. We have the activation memory of CoLA-M as
\begin{equation}
    M_{\text{CoLA-M}} = 2nd + 7nr.
\end{equation}
We summarize the results in Table~\ref{tab:compute-memory-gcp}.

\section{Training Configurations}
\label{sec:appendix-hyper-param}
\subsection{LLaMA Pre-Training}
\label{sec:appendix-llama-pt}
For optimizer related hyper-parameters, we empirically found 0.003 is a balanced choice of learning rate for most of the models we trained, this is similar to the settings in \cite{han2024sltrain}. For CoLA-1B, this learning rate triggers a unstable loss curve, thereby is reduced to 0.002, and is further reduced to 0.001 for CoLA-7B as a conservative practice. For smaller models like CoLA-60M, an even larger learning rate such 0.006 can be adopted. For the warm-up ratio, weight decay and gradient clipping, we found the commonly adopted settings, 0.1, 0.01, 0.5, are proper choices for CoLA. Other than the standard optimizer parameters, one needs to pre-define a rank $r$ when initializing CoLA. A default choice is set to approximately one quarter of the model inner width, i.e., $r=\frac{1}{4}d$. 

\subsection{BERT$_{\text{Large}}$ Pre-Training}
\label{sec:appendix-bert-pt}

\begin{table*}[t]
\centering
\small
\resizebox{\linewidth}{!}{%
\begin{tabular}{c|c|c|c|c|c|c|c|c|c|c}
\toprule
& Loss & QQP & SST-2 & MRPC & COLA & QNLI & MNLI & RTE & STS-B & GLUE Avg \\
\midrule
BERT$_{\text{Large}}$ & 1.263 & 91.1 & 92.1 & {90.7} & 53.1 & 91.6 & 84.3 & 69.9 & 88.9 & 82.7\\
\midrule
CoLA -- Gated MLP, Low-Rank \(\sigma\) Only & {\bf 1.257} & {\bf 91.2} & {\bf 92.3} & 90.6 & {54.1} & {\bf 91.7} & {\bf 84.3} & {\bf 74.2} & {89.7} & {\bf 83.5} \\
\midrule
CoLA -- Original MLP, Preserve Full-Rank \(\sigma\) & {1.265} & {\bf 91.2} & {92.1} & {\bf 91.7} & {\bf 55.1} & {91.5} & {83.7} & {73.1} & {\bf 89.8} & {\bf 83.5} \\
\bottomrule
\end{tabular}%
}
\caption{Fine-tuning CoLA and BERT$_{\text{Large}}$ on GLUE. Both models are trained from scratch following NVIDIA's faithful reproduction\footref{nv-rep}, then fine-tuned for three epochs. F1 scores are reported for MRPC, Pearson correlations are reported for STS-B, Matthews correlations are reported for COLA (task), accuracies are reported for all other tasks. Reported metrics are the mean of 5 best out of 10 random seeds. Two CoLA results are provides: "CoLA -- Gated MLP, Low-Rank \(\sigma\) Only" is the one shown in Table~\ref{tab:bert-ft}, in which the MLP structure is modified to have a gating module, so that it's viable to have only the low-rank activation; "CoLA -- Original MLP, Preserve Full-Rank \(\sigma\)" is an exact BERT architecture with all linear layers replaced by CoLA layers, so that both low-rank and full-rank activations exist.}
\label{tab:bert-ft-both-solutions}
\end{table*}

We directly adopted NVIDIA's open-sourced reproduction of BERT pre-training\footref{nv-rep}, without changing any training configurations or hyper-parameters (including learning rate). We implemented CoLA onto this training pipeline and set CoLA as $0.7\times$ compute of full-rank BERT$_{\text{Large}}$, which corresponds to rank 384 at attention layers and rank 512 at MLP layers. We choose this setting due to its superior performance observed in Table~\ref{tab:cola-scaling-up}.

Both CoLA and BERT$_{\text{Large}}$ are trained for 85B tokens using masked token prediction and next sentence prediction, with a composition of 128 tokens per sequence in 90\% steps and 512 tokens per sequence in the rest 10\% steps. Most settings in this reproduction are identical to the original BERT \cite{devlin2019bert}, except the adoption of LAMB optimizer \cite{you2019large} for large batch training and the constraint of using only the Wikipedia corpus. We kept everything unchanged, and successfully reproduced BERT$_{\text{Large}}$ as training loss of 1.263, very close to the mean value 1.265 reported by NVIDIA. Meanwhile, we trained CoLA using the exact same training configurations and got the training loss of {\bf 1.257}, suggesting a slightly better outcome despite of fewer parameter and compute.

The only caveat of adopting CoLA onto BERT is, we can't remove the full-rank activation unless we modify its MLP structure, because the original BERT MLP has a two-layer structure, i.e., \(\mat{W}_2\sigma(\mat{W}_1\mat{x)}\). If we adopt CoLA while removing the full-rank activation, it becomes \(\mat{B}_2\sigma\left(\mat{A}_2\mat{B}_1\sigma(\mat{A}_1\mat{x})\right)\), in which \(\mat{A}_2\) and \(\mat{B}_1\) are adjacent with no other operations in between, therefore \(\mat{A}_2\mat{B}_1\) is mathematically equivalent to an \(r\) to \(r\) linear transformation. This could lead to a significant performance drop: we tried this setup at phase 1 (sequence length 128), resulting in a higher pre-training loss 1.579 vs 1.403. To avoid this setup, we naturally have two solutions: (1) use gated MLP, such as the ones in LLaMA/Mixtral/Qwen Models, then the full-rank activation can be safely removed; (2) use the original non-gated MLP, but preserve the full-rank activation on top of CoLA. In both solutions, a non-linear operation is placed between \(\mat{A}_2\) and \(\mat{B}_1\). We clarify that results shown in Section~\ref{sec:results} are using solution (1).

To isolate the effect of CoLA from the structural change of the MLP layer, we also show results from solution (2) and compare with those from solution (1) side by side in Table~\ref{tab:bert-ft-both-solutions}. Interestingly, the average score for both solutions happen to be the same, with some variations in task-level performance. At similar performance, solution (1) yields lower FLOPs and memory cost, and is more aligned with CoLA's design principal. Therefore, we promote solution (1).

\section{Additional Results}

\subsection{Ablation Study}
\begin{table}[t]
\centering
\small
\begin{tabular}{c|c|c|c}
\toprule
& 60M & 130M & 350M \\
\midrule
CoLA w/ Both $\sigma$ & {\bf 34.04} & {\bf 24.48} & 19.56 \\
\midrule
CoLA w/ Only Low-Rank $\sigma$ & 34.35 & 25.20 & {\bf 19.40} \\
\midrule
\makecell[c]{CoLA w/ Only Low-Rank $\sigma$ \\ -- Reduced} & 35.41 & 25.90 & 20.50 \\
\midrule
CoLA w/ Only Full-Rank $\sigma$ & 36.26 & 26.85 & 21.18 \\
\bottomrule
\end{tabular}%
\caption{Ablation study regarding where to place the low-rank non-linear functions.}
\label{tab:result-ablation}
\end{table}

\label{sec:appendix-ablation}
We empirically found that keeping the original LLaMA nonlinearity on top of our proposed formulation Eq.~\eqref{eq:main-fwd} helps improve the model performance at smaller scales, such as 60M and 130M. However, when scaling up to 350M we no longer observe such a benefit. Therefore, the default setting of pre-training CoLA-1B/7B is set to use only low-rank nonlinearity. We found also evident that applying low-rank nonlinearity (i.e., Eq.~\eqref{eq:main-fwd}) regardless of whether the original linear layer being followed by nonlinearity is crucial to boost model performance. Results are shown in Table.~\ref{tab:result-ablation}, in which "CoLA w/ Both $\sigma$" means keeping the original nonlinearity on top of proposed low-rank nonlinearity, "CoLA w/ Only Low-Rank $\sigma$" means applying Eq.~\eqref{eq:main-fwd} in an agnostic way to all linear layers, "CoLA w/ Only Low-Rank $\sigma$ -- Reduced" means only applying Eq.~\eqref{eq:main-fwd} to the linear layers that are originally followed by nonlinearity, "CoLA w/ Only Full-Rank $\sigma$" means keeping the low-rank factorization but does not apply low-rank nonlinearity.

\subsection{Inference Efficiency}
\label{sec:inference}
\begin{table}[t]
\centering
\small
\resizebox{0.85\linewidth}{!}{%
\begin{tabular}{c|cc|cc}
\toprule
& \multicolumn{2}{c|}{1B (BZ=32)} & 
\multicolumn{2}{c}{7B (BZ=32)} \\
\cmidrule{2-5}
& Mem (GB) & Token/s & Mem (GB) & Token/s\\
\midrule
Full-rank & 5.74 & 21,109 & 18.15 & 11,086 \\
\midrule
SLTrain & 4.18 & 20,096 & 12.70 & 9,968 \\
\midrule
CoLA & {\bf 3.84} & {\bf 34,697} & {\bf 10.87} & {\bf 16,012} \\
\bottomrule
\end{tabular}%
}
\caption{Comparison of memory (GB) and throughput (Token/sec) at inference time on an A100 GPU.}
\label{tab:inference}
\end{table}

We show CoLA's system performance at inference stage in Table~\ref{tab:inference}. CoLA reduces memory usage and improves inference throughput compared to full-rank baselines.

\section{Detailed Profiling Setting}
\label{sec:appendix-detailed-experiment-setting}
This section provides a detailed explanation of the experimental setup for system-level measurements. For the memory breakdown in Fig.~\ref{fig:gcp-memory-break}, we use a sequence batch size of 32. For throughput measurement in Fig.~\ref{fig:throughput}, we use a sequence batch size of 16 because the full-rank model cannot fit into 40GB A100 when using a sequence batch size of 32. Throughput is measured incorporating one forward pass, one backward pass, and one optimizer step. This setup reflects a realistic training scenario, particularly in a multi-GPU environment, such as an 8x A100 cluster utilizing simple data parallelism. For a fair comparison, we set the update step in GaLore/APOLLO to 200, ensuring that the computationally expensive SVD/random projection is performed only once every 200 optimizer steps and is distributed across a single optimizer step. All experiments are conducted on a single GPU to isolate the effected of FLOP reduction on throughput improvement, without being influenced by multi-GPU framework settings or communication overhead. For Table.~\ref{tab:result-7b}, memory consumption is measured on a 94GB H100 with a sequence batch size of 16. For Table.~\ref{tab:inference}, inference is performed using the same configuration as pre-training, with a sequence batch size of 32.

\section{Proof of Theoretical Results}\label{section: proof of main results}
\begin{proof}[Proof of Proposition \ref{prop: erho <= eone}]
    Note that the activation function $\sigma$ applies element-wisely. With the assumption of $\sigma(0)=0$, by Taylor's expansion of $\sigma$ at $0$, it holds that for any $\mat{Z}\in\mathbb{R}^{r\times n}$, 
    \begin{equation}\label{in the proof: taylor expansion of rho}
        \sigma(\tau \mat{Z})=\sigma'(0)\tau \mat{Z} + R(\tau,\mat{Z}),
    \end{equation}
    where $R(\tau,\mat{Z})$ is matrix-valued function satisfying 
    \begin{equation}\label{eq: lim norm max of R / tau = 0}
        \lim_{\tau\to0}\frac{\norm{R(\tau,\mat{Z})}_{\max}}{|\tau|}=0.
    \end{equation}
    By $(\mat{A}^*,\mat{B}^*)$ we denote the optimal solution of $\Eone(r)$. With assumption of $\sigma'(0)\neq0$, we let $\mat{A_\tau:=\tau A^*}$ and $\mat{B}_\tau:=\frac{1}{\sigma'(0)\tau}\mat{B}^*$, for $\tau\neq0$. It follows from \eqref{in the proof: taylor expansion of rho} with $\mat{Z=A^*X}$ that 
    \begin{align*}
    \mat{B}_\tau\sigma(\mat{A}_\tau\mat{X})&=\mat{B}^*\mat{A}^*\mat{X} + \mat{B}^*\frac{R(\tau,\mat{A^*X})}{\tau}\frac{1}{\sigma'(0)}. 
    \end{align*}
    Note that 
    \begin{align*}
&\normF{\mat{B}^*\frac{R(\tau,\mat{A^*X})}{\tau}\frac{1}{\sigma'(0)}}\\
&\leq \normTwo{\mat{B}^*}\sqrt{rn}\frac{\norm{R(\tau,\mat{A^*X})}_{\max}}{|\tau|}\frac{1}{|\sigma'(0)|}.
    \end{align*}
    which combining with the property \eqref{eq: lim norm max of R / tau = 0} with $\mat{Z=A^*X}$ indicates  
    \begin{equation*}
\lim_{\tau\to0}\normF{\mat{B}^*\frac{R(\tau,\mat{A^*X})}{\tau}\frac{1}{\sigma'(0)}}=0.
    \end{equation*}
    This is equivalent to say $$\lim_{\tau\to0}\normF{\mat{B}_\tau\sigma(\mat{A}_\tau\mat{X})-\mat{B^*A^*X}}=0,$$
    which implies 
    \begin{equation}\label{in the proof: lim tau to 0 normF XW - rho(XAtau)Btau}
    \lim_{\tau\to0}\normF{\mat{Y}-\mat{B}_\tau\sigma(\mat{A}_\tau\mat{X})}=\underbrace{\normF{\mat{Y}-\mat{B}^*\mat{A}^*\mat{X}}}_{=\Eone(r)}. 
    \end{equation}
    Note that for any $\tau\neq0$, there holds $\Esigma(r)\leq \normF{\mat{Y}-\mat{B}_\tau\sigma(\mat{A}_\tau\mat{X})}$. This combining with \eqref{in the proof: lim tau to 0 normF XW - rho(XAtau)Btau} yields $\Esigma(r)\leq \Eone(r)$ as desired. 
\end{proof}

\begin{proof}[Proof of Proposition \ref{prop: rho Xu not in col(X)}]
    According to Lemma \ref{lemma: Xtop diag(w) X neq 0}, there exists 
    \begin{equation}\label{in the proof: w in ker XT}
    \mat{w}\in\ker(\mat{X})\backslash\{\mat{0}\},
    \end{equation}
    such that 
    \begin{equation}\label{in the proof: Xtop diagw X neq 0}
    \mat{X}\mathrm{diag}(\mat{w})\mat{X}^\top\neq \mat{0}. 
    \end{equation}
    We define a function $g:\mathbb{R}^{\din }\to\mathbb{R}$ by $g(\mat{u}):=\sigma(\mat{u^\top X})\mat{w}$. Direct computation gives $\nabla g(\mat{u})=\mat{X}\parens{\mat{w\odot\sigma'(X^\top u)}}$ and $\nabla^2g(\mat{u})=\mat{X}\mathrm{diag}(\mat{w}\odot\sigma''(\mat{X^\top u}))\mat{X}^\top$. By assumptions of $\sigma(0)=0$, $\sigma'(0)\neq0$ and $\sigma''(0)\neq0$, we have $g(\mat{0})=0$, $\nabla g(\mat{0})=\sigma'(0)\mat{X} \mat{w}=\mat{0}$ (due to \eqref{in the proof: w in ker XT}) and $\nabla^2g(\mat{0})=\sigma''(0)\mat{X}\mathrm{diag}(\mat{w})\mat{X}^\top\neq\mat{0}$ (due to \eqref{in the proof: Xtop diagw X neq 0}). It results that $g$ is not the zero function. Therefore, there exists $\mat{u}\in\mathbb{R}^{\din }$ such that $g(\mat{u})\neq 0$, that is $\sigma(\mat{u^\top X})\mat{w}\neq 0$ (hence $\sigma(\mat{u^\top X})$ is a nonzero vector). Since $\mat{w}\in\ker(\mat{X})\backslash\{\mat{0}\}$, we know that $\sigma(\mat{X^\top u})\notin \parens{\ker(\mat{X})}^\perp=\col(\mat{X^\top})$, or equivalently $\sigma(\mat{u^\top X})\notin\row(\mat{X})$. This completes the proof. 
\end{proof}

\paragraph{Discussion.} 
The proof of Proposition \ref{prop: rho Xu not in col(X)} is constructive. The function $g$ constructed therein is continuous and not identically zero. We mention that any $\mat{u}$ in the support set $\{\mat{u}:g(\mat{u})\neq0\}$ will satisfy $\sigma(\mat{u}^\top \mat{X})\notin \row(\mat{X})$. 

\begin{proof}[Proof of Theorem \ref{theorem: Erho < Eid}]
     Noting that $\mat{Y}=\mat{Y}_\perp+\mat{Y}_\parallel$; rows of $\mat{Y}_\perp$ are in $\row(\mat{X})^\perp$; and rows of $\mat{Y}_\parallel$ and $\mat{BAX}$ are in $\row(\mat{X})$, we have $\normF{\mat{Y-BAX}}^2=\normF{\mat{Y}_\perp}^2+\normF{\mat{Y}_\parallel-\mat{BAX}}^2.$ 
     Since rows of $\mat{Y}_\parallel$ belong to $\row(\mat{X})$, there exists $\mat{W}\in\mathbb{R}^{\dout\times \din}$ such that $\mat{Y}_\parallel=\mat{WX}$. Then it follows from Lemma \ref{lemma: Eone=Etwo=Ethree} that $\Eone(r)^2=\normF{\mat{Y}_\perp}^2+\parens{s_{>r}(\mat{Y}_\parallel)}^2$.
    Now we consider $\Esigma(r)$. By triangle inequality, for any $\mat{A}\in\mathbb{R}^{r\times\din}$ and $\mat{B}\in\mathbb{R}^{\dout\times r}$, we have $\normF{\mat{Y-B\sigma(AX)}}\leq \normF{P_{\mat{v}}(\mat{Y})-\mat{B\sigma(AX)}}+\normF{P_{\mat{v}^\perp}(\mat{Y})}$. Since orthogonal projection is done row-wisely, there exists $\mat{w}\in\mathbb{R}^{\dout}$ such that $P_{\mat{v}}(\mat{Y})=\mat{wv^\top}$. Then according to Lemma \ref{lemma: Esigma=0}, we have 
    \begin{equation*}
    \min_{\mat{A}\in\mathbb{R}^{r\times \din}, \mat{B}\in\mathbb{R}^{\dout\times r}}\normF{P_{\mat{v}}(\mat{Y})-\mat{B\sigma(AX)}}=0. 
    \end{equation*}
    Therefore, $\Esigma(r)\leq \normF{P_{\mat{v}^\perp}(\mat{Y})}$. Then the desired result immediately follows from assumption \eqref{eq: assumption strictly smaller}. 
\end{proof}

\paragraph{Discussion.} An instructive extreme case in Theorem \ref{theorem: Erho < Eid} is when every row of $\mat{Y}$ lies in $\mathrm{span}\{\mat{v}^\top\}$. Then the left side of \eqref{eq: assumption strictly smaller} is $0$, while the right side is
strictly positive since $\mat{v^\top\notin\row(X)}$ implies $\normF{\mat{Y}_\perp}\neq0$. Hence, assumption \eqref{eq: assumption strictly smaller} clearly holds. Moreover, as indicated in the proof of Theorem \ref{theorem: Erho < Eid}, $(\Esigma(r))^2$ is bounded above by the left side of \eqref{eq: assumption strictly smaller} and $(\Eone(r))^2$ is equal to the right side. Therefore, $0=\Esigma(r)<\Eone(r)$ in this specific case. 

\begin{proof}[Proof of Theorem \ref{theorem: effective rank}]
It follows from triangle inequality that 
\begin{align*}
\Delta\leq \normF{\mat{B_{\mathrm{True}}\sigma(A_{\mathrm{True}}X)-Y}}+\Esigma(r).
\end{align*}
Denote $k:=r_\alpha(\mat{Y})$. Let $s_i$ be the singular values of matrix $\mat{Y}$ in a non-increasing order. Let \(\mat{Y}_1\) be the rank-\(k\) truncated singular value decomposition of $\mat{Y}$ (keeping the top \(k\) singular values and setting the rest to zero), and
\(\mat{Y}_2:=\mat{Y}-\mat{Y}_1\) be the residual (zeroing the top \(k\) singular
values and keeping the remaining). It results that
\begin{equation}\label{in the proof: rank of Y1}
    \rank(\mat{Y}_1)=k,
\end{equation}
and
\begin{equation}\label{in the proof: property of Y2}
    \normTwo{\mat{Y}_2}=s_{k+1},\quad \normF{\mat{Y}_2}=s_{>k}(\mat{Y}). 
\end{equation}
Now
\begin{align*}
&\normF{\mat{B_{\mathrm{True}}\sigma(A_{\mathrm{True}}X)-Y}}\\
\leq&\normF{\mat{B_{\mathrm{True}}\sigma(A_{\mathrm{True}}X)}-\mat{Y}_1}+\normF{\mat{Y}_2}\stepjust{triangle inequality}\\
\leq&\sqrt{r+k}\normTwo{\mat{B_{\mathrm{True}}\sigma(A_{\mathrm{True}}X)}-\mat{Y}_1}+\normF{\mat{Y}_2}\stepjust{Lemma \ref{lemma: F norm leq 2 norm} and Eq. \eqref{in the proof: rank of Y1}}\\
\leq&\sqrt{r+k}\normTwo{\mat{B_{\mathrm{True}}\sigma(A_{\mathrm{True}}X)-Y}}\\
&\qquad+\sqrt{r+k}\normTwo{\mat{Y}_2}+\normF{\mat{Y}_2}\stepjust{triangle inequality}\\
\leq&\sqrt{r+k}\parens{\normTwo{\mat{G}}+\epsilon} +s_{k+1}\sqrt{r+k}+s_{>k}(\mat{Y}).\stepjust{assumption \eqref{eq: assumption XW - rhoBAX - G} and Eq. \eqref{in the proof: property of Y2}}
\end{align*}
According to Theorem 4.6.1 in \citet{vershynin2018high}, with probability at least $1-2\exp(-(n+\dout))$, $\normTwo{\mat{G}} \leq C v\sqrt{n+\dout}$.
The desired result immediately follows by substituting the estimate of $\normTwo{\mat{G}}$. 
\end{proof}
\paragraph{Discussion.}
We remark that, setting $\alpha=1$, the error bound in Theorem~\ref{theorem: effective rank} reduces to the full-rank case 
$\sqrt{r+\rank(\mat{Y})}\parens{Cv\sqrt{n+\dout}+\epsilon}+\Esigma(r)$, 
since $r_1(\mat{Y})=\rank(\mat{Y})$, $s_{r_1(\mat{Y})+1}=0$, and $s_{>r_1(\mat{Y})}(\mat{Y})=0$.

\section{Auxiliary Lemmas}
\begin{lemma}\label{lemma: Xtop diag(w) X neq 0}
    Suppose that matrix $\mat{X}\in\mathbb{R}^{\din\times n}$ has no identical columns, no zero columns and satisfies $n>\rank(\mat{X})$. Then there exists $\mat{w}\in\ker(\mat{X})\backslash\{\mat{0}\}$ such that 
    \begin{equation}\label{eq: Xtop diagw X neq 0}
    \mat{X}\mathrm{diag}(\mat{w})\mat{X}^\top\neq \mat{0}. 
    \end{equation}
\end{lemma}

\begin{proof}
    Note that the assumption of $n>\rank(\mat{X})$ guarantees $\ker(\mat{X})$ is non-trivial. We pick an element $\mat{w}=[w_i:i\in[n]]\in\ker(\mat{X})\backslash\{\mat{0}\}$ such that $\mat{w}$ has a minimum number of nonzero entries. We denote the support of $\mat{w}$ by $S:=\{i:w_i\neq0\}$. We remark that $|S|$ is in fact the \textit{spark} of matrix $\mat{X}$, which is the smallest number of columns of $\mat{X}$ that are linearly dependent. By $\mat{x}_i$ we denote the $i$-th column vector of $\mat{X}$ for each $i\in[n]$. The definition of $\mat{w}$ implies
    \begin{equation}\label{in the proof: sum wixi}
        \sum_{i\in S}w_i\mat{x}_i=0, 
    \end{equation}
    and elements in any proper subset of $\{\mat{x}_i:i\in S\}$ are linearly independent. 
    
    Since $\mat{X}$ has no zero columns, we know that $|S|\geq2$. If $|S|=2$, without loss of generality, we assume that $S=\{1,2\}$. Then \eqref{in the proof: sum wixi} gives $w_1\mat{x}_1+w_2\mat{x}_2=0$. It follows that $\mat{X}\mathrm{diag}(\mat{w})\mat{X}^\top=w_1\mat{x}_1 \mat{x}_1^\top+w_2\mat{x}_2 \mat{x}_2^\top=\frac{w_1(w_1+w_2)}{w_2}\mat{x}_1 \mat{x}_1^\top$. 
    Noting that $w_1\neq0$, $w_2\neq0$, $w_1+w_2\neq0$ (otherwise $\mat{x}_1=\mat{x}_2$ contradicting to assumption of no identical columns) and $\mat{x}_1 \mat{x}_1^\top\neq\mat{0}$ (otherwise $\mat{x}_1=\mat{0}$ contradicting to assumption of no zero columns), therefore \eqref{eq: Xtop diagw X neq 0} holds.  

    If $|S|\geq3$, without loss of generality, we assume that $S=[k]$ with $k\geq3$. Let 
    $S^-:=S\backslash\{1\}$. Note that $\mat{x}_1\in\mathcal{X}:=\mathrm{span}\{\mat{x}_i:i\in S^-\}$ and $\{\mat{x}_i:i\in S^-\}$ are linearly independent. Then it is clear that there exists a nonzero vector $\mat{y}\in \mathcal{X}$ such that $\mat{x}_1^\top\mat{y}=0$. Since $\mat{y}$ is a nonzero vector in $\mathcal{X}$, there exists $j_0\in S^-$ such that $\mat{x}_{j_0}^\top\mat{y}\neq0$. Now we assume by contradiction that \eqref{eq: Xtop diagw X neq 0} does not hold, that is, $\mat{X}\mathrm{diag}(\mat{w})\mat{X}^\top$ is a zero matrix. Then we have $\mat{X}\mathrm{diag}(\mat{w})\mat{X}^\top\mat{y}=\mat{0}$, or entry-wisely $\sum_{i\in S} w_i(\mat{x}_i^\top\mat{y})\mat{x}_i=\mat{0}$. By definition of $\mat{y}$, we have $\mat{x}_1^\top\mat{y}=0$, which implies $\sum_{i\in S^-} w_i(\mat{x}_i^\top \mat{y})\mat{x}_i=0$. However, noting that $\mat{x}_{j_0}^\top\mat{y}\neq 0$ for $j_0\in S^-$, the above equation contradicts to the fact that $\{\mat{x}_i:i\in S^-\}$ are linearly independent. Therefore, we conclude by contradiction that \eqref{eq: Xtop diagw X neq 0} holds. 
\end{proof}

\begin{lemma}\label{lemma: Esigma=0}
    Suppose that $\mat{u}\in\mathbb{R}^{\din}$ and $\mat{v^\top:=\sigma(u^\top X)\notin\row(X)}$. If there exists $\mat{w}\in\mathbb{R}^{\dout}$ such that $\mat{Y=wv^\top}$, then $\Esigma(r)=0$. 
\end{lemma}
\begin{proof}
    Let $\mat{A^*}\in\mathbb{R}^{r\times\din}$ be the matrix whose each row is $\mat{u}^\top$. And let $\mat{B^*}\in\mathbb{R}^{\dout\times r}$ be the matrix whose first column is $\mat{w}$ and the other columns are zeros. Direct computation gives $\mat{B^*\sigma(A^*X)}=\mat{wv^\top}$ which is exactly $\mat{Y}$. Therefore, $\Esigma(r)=0$. 
\end{proof}

\begin{lemma}\label{lemma: Eone=Etwo=Ethree}
Suppose that there exists $\mat{W}\in\mathbb{R}^{\dout\times\din}$ such that $\mat{Y=WX}$. Then it holds that $\Eone(r)=s_{>r}(\mat{WX}).$
\end{lemma}

\begin{proof}
Define 
\begin{align*}
\Etwo(r)&:=\min_{\widetilde{\mat{W}}\in\bR^{\dout\times \din},\rank(\widetilde{\mat{W}})\leq r}\normF{\mat{WX-\widetilde{W}X}},\\
\Ethree(r)&:=\min_{\mat{M}\in\bR^{\dout\times n},\rank(\mat{M})\leq r}\normF{\mat{WX-M}}.
\end{align*} 
Since $\rank(\mat{BAX})\leq r$ and $\rank(\mat{\widetilde{W}X})\leq r$, it is clear that $\Ethree(r)\leq \Eone(r)$ and $\Ethree(r)\leq \Etwo(r)$. Note that by Eckart–Young–Mirsky theorem, the optimal solution of $\Ethree(r)$, denoted by $\mat{M}^*$, is a rank $r$ matrix obtained by the truncated singular value decomposition of $\mat{WX}$; moreover, $\Eone(r)=s_{>r}(\mat{WX})$. Let $\mat{X}^\dagger$ be the pseudoinverse of $\mat{X}$. Since $\row(\mat{M}^*)\subset\row(\mat{WX})\subset\row(\mat{X})$ and $\mat{X^\dagger X}$ is the orthogonal projection onto $\row(\mat{X})$, we have $\mat{M^*X^\dagger X=M^*}$. Note that $\rank(\mat{M}^*\mat{X}^\dagger)\leq\rank(\mat{M}^*)\leq r$. Therefore, $\Etwo(r)\leq\normF{\mat{WX-M^*X^\dagger X}}=\normF{\mat{WX-M^*}}=\Ethree(r).$ Above all, we have $\Etwo(r)=\Ethree(r)$. Since $\rank(\mat{M^* X^\dagger})\leq r$, with singular value decomposition on $\mat{M^*X^\dagger}$, one can always find $\mat{A}^*\in\mathbb{R}^{r\times\din}$ and $\mat{B}^*\in\mathbb{R}^{\dout\times r}$ such that $\mat{ M^*X^\dagger=B^*A^*}$. This implies $\Eone(r)\leq \Ethree(r)$. Hence, we obtain $\Eone(r)=\Etwo(r)=\Ethree(r)=s_{>r}(\mat{WX})$ as desired. 
\end{proof}

\begin{lemma}\label{lemma: F norm leq 2 norm}
For any matrices $\mat{C}$ and $\mat{D}$, it holds that $\normF{\mat{C-D}}\leq \sqrt{\rank(\mat{C})+\rank(\mat{D})}\normTwo{\mat{C-D}}$.
\end{lemma}
\begin{proof}
    Let $s_i$ be the singular values of matrix $\mat{C-D}$ in a non-increasing order.
    Note that $\rank(\mat{C-D})\leq\rank(\mat{C})+\rank(\mat{D})$. Then $\normF{\mat{C-D}}^2=\sum_i s_i^2\leq \rank(\mat{C-D}) s_1^2\leq \parens{\rank(\mat{C})+\rank(\mat{D})}\normTwo{\mat{C-D}}^2$. 
\end{proof}
\end{document}